\documentclass[letterpaper]{article} % DO NOT CHANGE THIS
\usepackage{aaai23}  % DO NOT CHANGE THIS
\usepackage{times}  % DO NOT CHANGE THIS
\usepackage{helvet}  % DO NOT CHANGE THIS
\usepackage{courier}  % DO NOT CHANGE THIS
\usepackage[hyphens]{url}  % DO NOT CHANGE THIS
\usepackage{graphicx} % DO NOT CHANGE THIS
\usepackage{natbib}  % DO NOT CHANGE THIS AND DO NOT ADD ANY OPTIONS TO IT
\usepackage{caption} % DO NOT CHANGE THIS AND DO NOT ADD ANY OPTIONS TO IT
\usepackage{algorithm}
\usepackage{algorithmic}

\usepackage{newfloat}
\usepackage{listings}
\DeclareCaptionStyle{ruled}{labelfont=normalfont,labelsep=colon,strut=off} % DO NOT CHANGE THIS
\floatstyle{ruled}
\newfloat{listing}{tb}{lst}{}
\floatname{listing}{Listing}
\usepackage{paralist}
\usepackage[utf8]{inputenc} % allow utf-8 input
\usepackage[T1]{fontenc}    % use 8-bit T1 fonts
\usepackage{url}            % simple URL typesetting
\usepackage{booktabs}       % professional-quality tables
\usepackage{amsfonts}       % blackboard math symbols
\usepackage{nicefrac}       % compact symbols for 1/2, etc.
\usepackage{microtype}      % microtypography

\usepackage{csquotes}
\usepackage{amsmath}
\usepackage{amssymb}
\usepackage{placeins}
\usepackage{subcaption} % subfigure
\usepackage[dvipsnames]{xcolor}
\title{Combining Slow and Fast: \\Complementary Filtering for Dynamics Learning}
\author{Katharina Ensinger\textsuperscript{\rm 1,\rm 2}, Sebastian Ziesche\textsuperscript{\rm 1}, Barbara Rakitsch\textsuperscript{\rm 1}, \\
	Michael Tiemann\textsuperscript{\rm 1}, Sebastian Trimpe\textsuperscript{\rm 2} }
\affiliations{
    \textsuperscript{\rm 1} Bosch Center for Artificial Intelligence, Renningen, Germany\\
    \textsuperscript{\rm 2} Institute for Data Science in Mechanical Engineering, RWTH Aachen University

    katharina.ensinger@bosch.com, firstname.surname@de.bosch.com, trimpe@dsme.rwth-aachen.de
}

\usepackage{bibentry}
\begin{document}

\maketitle

\begin{abstract}
	Modeling an unknown dynamical system is crucial in order to predict the future behavior of the system. 
	A standard approach is training recurrent models on measurement data.
	While these models typically provide exact short-term predictions, accumulating errors yield deteriorated long-term behavior. 
	In contrast, models with reliable long-term predictions can often be obtained, either by training a robust but less detailed model, or by leveraging physics-based simulations.
	In both cases, inaccuracies in the models yield a lack of short-time details. 
	Thus, different models with contrastive properties on different time horizons are available. 
	This observation immediately raises the question: \emph{Can we obtain predictions that combine the best of both worlds?}
	Inspired by sensor fusion tasks, we interpret the problem in the frequency domain and leverage classical methods from signal processing, in particular complementary filters.
	This filtering technique combines two signals by applying a high-pass filter to one signal, and low-pass filtering the other. 
	Essentially, the high-pass filter extracts high-frequencies, whereas the low-pass filter extracts low frequencies. 
	Applying this concept to dynamics model learning enables the construction of models that yield accurate long- and short-term predictions. 
	Here, we propose two methods, one being purely learning-based and the other one being a hybrid model that requires an additional physics-based simulator.
\end{abstract}
% !TEX root = neurips_2021.tex

\section{Introduction} \label{section:intro}
%\begin{itemize}
%\item define and explain time-series data	
%\item define and explain simulator model 
%\item setup of data-based component
%\item intuition frequency domain
%\item complementary filter
%\item what do we achieve? 
%\end{itemize}
Many physical processes $\left(x_n\right)_{n=0}^N$ with $x_n \in \mathbb{R}^{D_x}$ can be described via a discrete-time dynamical system 
\begin{equation}\label{eq:dyn}
\begin{aligned}
x_{n+1}= f(x_n).
\end{aligned}
\end{equation}
Typically, it is not possible to measure the whole state-space of the system \eqref{eq:dyn}, but a function of the states corrupted by noise $\hat{y}_n$ can, for example, be measured by sensors 
\begin{equation}\label{eq:obs}
\begin{aligned}
y_n &= g(x_n)  = Cx_n, \\
\hat{y}_n  &=  y_n+\epsilon_n, \textrm{ with } \epsilon_n \sim \mathcal{N}(0,\sigma^2)
\end{aligned}
\end{equation}
and $C \in \mathbb{R}^{D_y \times D_x}$. 
Our general interest is to make accurate predictions for the observable components $y_n$ in Eq. \eqref{eq:obs}. 
One possible way to address this problem is training a recurrent model on the noisy measurements $\hat{y}_n$ in Eq. \eqref{eq:obs}. 
Learning-based methods are often able to accurately reflect the system's behavior and therefore produce accurate short-term predictions. 
However, the errors accumulate over time leading to deteriorated long-term behavior \citep{DBLP:conf/iclr/ZhouLXHH018}.  

To obtain reliable prediction behavior on each time scale, we propose to decompose the problem into two components. In particular, we aim to combine two separate models, where one component reliably predicts the long-term behavior, while the other adds short-term details, thus combining the strengths of each component. 
Interpreted in the frequency domain, one model tackles the low-frequency components while the other tackles the high-frequency parts. 

Combining high and low-frequency information from different signals or models is well-known from control engineering or signal processing tasks. 
One typical example is tilt estimation in robotics, where accelerometer and gyroscope data are often available simultaneously \citep{5509756, 9834094}. 
On one hand, the gyroscope provides position estimates that are precise on the short-term but due to integration in each time step, accumulating errors cause a drift on the long-term. 
On the other hand, the accelerometer-based position estimates are long-term stable, but considerably noisy and thus not reliable on the short-term. 
Interpreted in the frequency domain, the gyroscope is more reliable on high frequencies, whereas the accelerometer is more reliable on low frequencies. 
Therefore, a high-pass filter is applied to the gyroscope measurements, whereas a low-pass filter is applied to the accelerometer measurements. Both filtered components are subsequently combined in a new complementary filtered signal that is able to approximate the actual position more accurately.

Here, we adopt the concept of complementary filter pairs to our task to fuse models with contrastive properties. 
In general, a complementary filter pair consists of a high-pass filter $H$ and a low-pass filter $L$, where the filters map signals to signals.
Depending on the specific filter, certain frequencies are eliminated while others pass.  
Intuitively the joint information of both filters in a complementary filter pair covers the whole frequency domain.
Thus, the key concept that we leverage here is the decomposition of a signal $y=(y_n)_{n=0}^N$ into a high-pass filter component $H(y)$ and a low-pass filter component $L(y)$ via
\begin{equation}\label{eq:concept}
y=H(y)+L(y).
\end{equation}

\begin{figure*}[tb]
	\includegraphics[width= \textwidth]{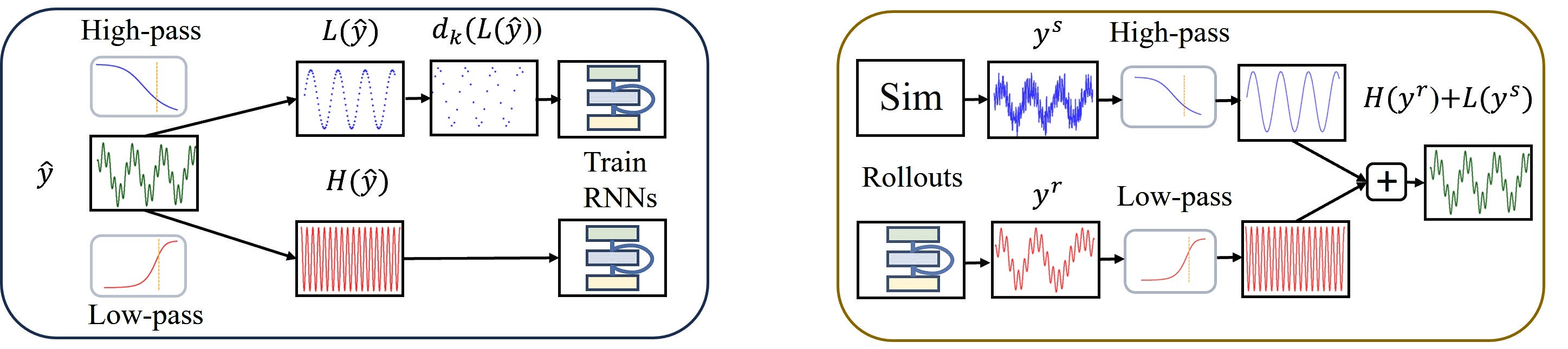}
	\caption{A high-level overview of our methods. Purely-learning-based scheme (left): a training signal is filtered into complementary components. The low-pass filtered signal is downsampled. Two seperate RNNs are trained on the decomposed signal. Hybrid model (right): The predictions of simulator and RNN are fed into the complementary filter. The resulting signal is trained end-to-end on the noisy observations by minimizing the root mean-squared error (RMSE). This structure is also applied to obtain predictions from the model.}
	\label{fig:scheme}
\end{figure*}

Based on the decomposition, we propose to address $H(y)$ and $L(y)$ by different models that are reliable on their specific time scale. 
\emph{In particular, we propose two methods, one being purely-learning based and one being a hybrid method that leverages an additional physics-based simulation.}
Both concepts are visualized in Figure \ref{fig:scheme}.
In the purely learning-based scenario, we train seperate networks that represent $H(y)$ and $L(y)$ in Eq. \eqref{eq:concept}.
In order to obtain a low-frequency model that indeed provides accurate long-term predictions, we apply a downsampling technique to the training data, thus reducing the number of integration steps. During inference, the predictions are upsampled up to the original sampling rate.
Applying the low-pass filter allows lossless downsampling of the signal depending on the downsampling ratio. 

In the hybrid scenario, only a single model is trained.  
Hybrid modeling addresses the problem of producing predictions by mixing different models that are either learning-based or obtained from first principles, e.g. physics \citep{yin2021augmenting,Suhartono_2017}. 
Here, we consider the case where access to predictions $y^{\text s}$ for the system \eqref{eq:dyn} is provided by a physics-based simulator. 
Additional insights, such as access to the simulator's latent space or differentiability are not given. 
While physics-based approaches are typically robust and provide reliable long-term behavior, incomplete knowledge of the underlying physics leads to short-term errors in the model.
Hence, we consider the case where $L(y^{\text s}) \approx L(y)$ holds.
By training a model for $H(y)$, the decomposition \eqref{eq:concept} becomes a hybrid model that combines the strengths of both components.   
The filter pair $(L,H)$ is integrated into the training process, assuring that the long-term behavior is indeed solely addressed by the simulator. 
In both scenarios, the learning-based and the hybrid, recurrent neural networks (RNNs) are trained on whole trajectories.  

In summary, the main contributions of this paper are:
\begin{compactitem}
	\item By leveraging complementary filters, we propose a new view on dynamics model learning;
	\item we propose a purely learning-based and a hybrid method that decompose the learning problem into a long-term and a short-term component; and
	\item we show that this decomposition allows for training models that provide accurate long and short-term predictions.
\end{compactitem}

%\input{problem_formulation} 
% !TEX root = neurips_2021.tex
\section{Related work} \label{section:rel}
%\begin{itemize}
%	\item Hybrid modelling (augmenting physical models, {\tiny Integrating Expert ODEs into Neural ODEs})
%	\item Latent force models
%	\item likelihood-free inference  
%	\item time-series modelling 
%\end{itemize}

In this section, we give an overview of related literature.

Several works point out parallels between classical signal-theoretic concepts and neural network architectures.
In particular, connections to finite-impulse response (FIR) and infinite-impulse response (IIR) filters have been drawn. 
The relations between these filters and feedforward-models have been investigated in \citet{6795541}.
Precisely, they construct different feedforward architectures by building synapses from different filters. 
Depending on the specific type, locally recurrent but globally feedforward structure can be obtained. 
These models are revisited in \citet{Campolucci} by introducing a novel backpropagation technique. 
More recently, feedforward Sequential Memory Networks are introduced, which can be interpreted as FIR filters \citep{zhang2016feedforward}. 
Relations between fully recurrent models and filters have been drawn as well.  
The hidden structure of many recurrent networks can be identified with classical filters. 
\citet{Kuznetsov2020DIFFERENTIABLEIF} point out the relation between Elman-Networks and filters and
introduce trainable IIR structures that are applied to sound signals in the experiments section.
Precisely, an Elman network can be interpreted as simple first-order IIR filter. 
In \citet{oliva2017statistical}, long-term dependencies are modeled via a moving average in the hidden units. 
Moving averages can again be interpreted as special FIR filters. 
\citet{Stepleton2018LowpassRN} recover long-term dependencies via a hidden structure of memory pools that consist of first-order IIR filters. 
However, none of this works leverages complementary filters in order to capture effects on multiple time scales.
Additionally, none of these approaches addresses hybrid dynamics models. 
\citet{doi:10.1177/0142331218755234, CERTIC2011419, Milic} combine learning techniques and in particular gradient-descent with complementary filters.
However, they consider the automatical adaption of the filter parameters.
In contrast, we leverage complementary filters for learning, in particular dynamics learning.

Filters manipulate signals on the frequency domain and thus address spectral properties. 
In \citet{Kutz2016DynamicMD} and \citet{Lange2021FromFT}, a signal is identified via spectral methods that are transformed into a linear model. 
Koopman theory is then leveraged to lift the system to the nonlinear space again. 
However, in our work, we use filters in order to separate the predictions on different time-horizons. 
Thus, in contrast to these works, our methods can be combined with different (recurrent) architectures and therefore allow for computing predictions via state-of-the-art techniques.

Combining physics-based simulators with learning-based models is an emerging trend. 
Hybrid models produce predictions by taking both models into account. Typically, the simulator is extended or parts of the simulator are replaced. 
There is a vast literature that deals with hybrid models for dynamical systems or time-series data. 
A traditional approach is learning the errors or residua of simulator predictions and data \citep{Forssell97combiningsemi-physical, Suhartono_2017}.
%In particular, the linear trend is learned via a linear regression model, while a nonlinear component is learned via a neural network. 
Another common approach in hybrid modeling is extending a physics-based dynamics model with neural ODEs \citep{yin2021augmenting, qian2021integrating}.
However, in contrast to our approach, these hybrid architectures do not explicitly exploit characteristics of the simulator, in particular the long-term behavior.
\citet{10.1785/0120170293} construct a hybrid model for the prediction of seismic behavior. 
Similar to our setting, they consider the case where a physics-based simulation provides reliable predictions for low frequencies, whereas lacking of a reliable model for high frequencies. However, the approach differs significantly from ours since a neural network is trained on a mapping from low to high frequencies. Furthermore, they do not consider dynamics models.
Therefore, it is unclear how to apply the approach to our problem setting.

\section{Background} \label{section:background}
In this section, we provide the necessary background on signal processing and filtering.
For a more detailed introduction, we refer the reader to \citet{oppenheim1999discrete}.

\subsection{Motivation} 
Filters are linear time-invariant systems that aim to extract specific frequency components from a signal. 
Standard types are high-pass and low-pass filters.
Low-pass filters extract low frequencies and attenuate high frequencies, whereas high-pass filters extract high frequencies and attenuate low frequencies. 
Frequencies that are allowed to pass are determined by a desired cutoff frequency. 
Further, additional specifications play a principal role in filter design, such as pass- and stop-band fluctuations and width of the transition band \citep{oppenheim1999discrete}.

Technically, a filter $F$ is a mapping in the time domain $F: l^\infty \to l^\infty: y \mapsto Z^{-1}(\mathcal F(Z(y)))$, where $\mathcal F: \mathbb C \to \mathbb C$ is the so-called transfer function in the frequency domain, $l^\infty$ is the signal space of bounded sequences and $Z: l^\infty \to \mathbb C$ is the well-known z-transform. Hence, a filter is obtained by designing a transfer function $\mathcal F$ in the frequency domain. For the type of filters considered here, the structure of $\mathcal F$ allows to directly compute $F(y)$ via a recurrence equation in the time domain (see the appendix for more details).
A typical application of filters is, for example, the denoising of signals. 
Noise adds a high-frequency component to the signal and can therefore be tackled by applying a low-pass filter.

\subsection{IIR-filter}
Typical filter types are finite-impulse response (FIR) and infinite-impulse response (IIR) \citep{oppenheim1999discrete}.
Here, we consider IIR filters. 
In contrast to FIR filters, IIR filters possess internal feedback.
%Typically, they require a smaller order than FIR-filters in order include the desired properties into the filters. 
Filtering a signal $y$ via an IIR-filter yields a recurrence equation for the filtered signal $\tilde{y}=(\tilde{y}_n)_{n=0}^N$ given by 
\begin{equation}\label{eq:rearange}
\tilde y_n=\frac{1}{a_0}\left(\sum_{k=1}^P a_k \tilde{y}_{n-k}+ \sum_{k=0}^P b_k y_{n-k}\right),
\end{equation}
where $P$ describes the filter order. 
The filter coefficients $a_k$ and $b_k$ are obtained from filter design with respect to the desired properties in the frequency domain. 
A detailed derivation is given in the appendix. 
There are different strategies to initialize the first $P$ values $\tilde{y}_0,\dots,\tilde{y}_{P-1}$ \citep{initialize, 492552}.
%We explain our strategy in Sec. \ref{section:method}. 

\subsection{Complementary filter pairs}\label{sec:complementary}
A complementary filter pair consists of a high-pass filter transfer function $\mathcal{H}$ and a low-pass filter transfer function $\mathcal{L}$ \citep{4101411}, chosen in a way that they cover the whole frequency domain, thus 
\begin{equation}\label{eq:decomposition}
y \approx L(y)+H(y)
\end{equation}
for any signal $y \in l^{\infty}$. 
Applying the complementary filter pair to two different signals $y^{\text{h}}$ and $y^{\text{l}}$ via $\tilde{y}=L(y^{\text{l}})+H(y^{\text{h}})$ directly yields a recurrence equation for the complementary filtered signal $\tilde{y}$ given by 
\begin{equation} \label{eq:IIR}
\tilde{y}_n = \frac{1}{a_0} \left(\sum_{k=1}^P a_k \tilde{y}_{n-k}+\sum_{k=0}^P b_k y^{\text{h}}_{n-k}+\sum_{k=0}^{P} \tilde{b}_k y^{\text{l}}_{n-k}\right),  
\end{equation}
where $a_k,b_k$ describe the high-pass filter parameters and $a_k,\tilde{b}_k$ describe the low-pass filter parameters. 
To obtain a joint recurrence equation, the filters are forced to share the parameters $a_k$. 
However, this can be done without loss of generality. 

\paragraph{Perfect complement} \label{section:perfect}
There are different strategies to express the decomposition \eqref{eq:decomposition} mathematically. 
One way is to construct the perfect complement in the frequency domain such that $\mathcal{H}+\mathcal{L}=1$ \citep{s21061937}.
Applying the perfect complementary filter to two identical signals $y^{\text{h}}=y^{\text{l}}$ results in the same signal as output.
For the IIR complementary filter \eqref{eq:IIR} this holds if $\tilde{b}_k=a_k-b_k$.
A detailed derivation is moved to the appendix. 
However, depending on the desired behavior of the filters, perfectly complementary filters are not always favorable.
Different approaches have been investigated in \citet{VAIDYANATHAN, inproceedings}.

\section{Method} \label{section:method}
We present two methods that leverage the idea of complementary filters for dynamics model learning in order to produce accurate short- and long-term predictions. 
Our first approach is applicable to general dynamics model learning, whereas our second approach is a hybrid modeling technique.
In the second case, access to trajectory data produced by a physics-based simulator is required. 
The key ingredient of both models is a complementary filter pair $(H,L)$ with parameters $a_k,b_k,\tilde{a}_k$ and $\tilde b_k$ (cf. Sec. \ref{sec:complementary}). 
While in the hybrid case reliable long-term predictions are already provided by the simulator, the long-term predictions have to be addressed by an additional model in the purely learning-based scenario.

\subsection{Recurrent dynamics model learning}
First, we give an overview of the recurrent dynamics model learning structure that serves as a backbone for our method. 
Here, we consider a recurrent multilayer perceptron (MLP) and a gated recurrent unit (GRU) model \citep{cho-etal-2014-learning}.
However, the method is not restricted to that choice and could be combined with other recurrent architectures such as \citet{HochSchm97, pmlr-v80-doerr18a}.  
Consider a trainable neural network transition function $f_{\theta}:\mathbb{R}^{D_h \times D_y} \rightarrow \mathbb{R}^{D_h}$ and a linear observation model $C_{\theta} \in \mathbb{R}^{D_y \times D_h}$.
Here, $\theta$ defines the trainable parameters and $h$ the latent states with corresponding latent dimension $D_h$. 
Predictions are computed via 
\begin{equation}\label{eq:RNN}
\begin{aligned}
h_{n+1} & = f_{\theta}(h_n,y_n) \\
y_n & = C_{\theta}h_n,
\end{aligned}
\end{equation}
where the initial hidden state $h_0$ can be obtained from the past trajectory by training a recognition model similar to \citet{pmlr-v80-doerr18a} or by performing a warmup phase.
Details are provided in the appendix. 
The mapping $F_{\theta}:\mathbb{R}^{D_h \times D_y}  \times \mathbb{N} \rightarrow \mathbb{R}^{D_y \times N}$ that computes an $N$-step rollout via Eq. \eqref{eq:RNN} reads
\begin{equation} \label{eq:trajectory}
F_{\theta}(h_0,y_0,N)=y_{0:N},
\end{equation} 
where $y_{0:N} \in \mathbb{R}^{D_y \times N}$ defines a trajectory with $N$ steps. 
\subsection{Purely learning-based model}
Next, we dive into the details of constructing complementary filter-based learning schemes and introduce our methods. 
In the purely learning-based scenario, two different models are trained, wherein one model addresses the high-frequency parts and the other addresses the low-frequency parts (see Figure \ref{fig:scheme} (left).)
To this end, the training signal $\hat{y}$ is decomposed into a high-frequency component  $H(\hat{y})$ and a low-frequency component $L(\hat{y})$  via the complementary filter pair (cf. Sec. \ref{sec:complementary}). The models are trained separately on the decomposition.
In order to obtain a model that indeed provides stable long-term behavior, the low-frequency training data is downsampled.
During inference, the predicted signal is upsampled again. 
Downsampling yields a model that performs less integration steps and thus, produces less error accumulation. 
As an additional advantage, backpropagation through less integration steps is computationally more efficient. 
Applying the low-pass filter allows lossless downsampling up to a specific ratio that is determined by the Nyquist frequency. 
Intuitively, only high-frequency information is removed that is addressed by the second network during training and inference. 
Details are provided in the appendix.
Splitting the training signal and training the models separately ensures that one model indeed addresses the low-frequency part of the signal and thus, the long-term behavior. 
End-to-end training on the other hand might yield deteriorated long-term behavior since it generally allows a single network to tackle both- short, and long-term behavior.

\paragraph{Up and Downsampling: }
The downsampling operation $d_k:\mathbb{R}^{D_y \times N} \rightarrow \mathbb{R}^{D_y \times  \lfloor N/k \rfloor}$ 
maps a signal to a lower resolution by considering every $k^{th}$ step of the signal via
\begin{equation} \label{eq:downsampling}
d_k(y_{0:N})=(y_0,y_k,\dots,y_{k  \lfloor N/k \rfloor}).
\end{equation}
The reverse upsampling operation $u_k:\mathbb{R}^{D_y \times N} \rightarrow \mathbb{R}^{D_y \times kN}$ maps a signal to a higher resolution by filling in the missing data without adding high-frequency artifacts to the signal. Mathematically, this corresponds to an interpolation problem \citep{oppenheim1999discrete}. Here, we consider lossless downsampling, where tolerable downsampling ratios are determined by the cutoff frequency of the low-pass filter. 

\paragraph{Training: }
Consider training data $\hat{y}_{0:N} \in \mathbb{R}^{D_y \times N}$ from which the first $R<N$ steps $\hat{y}_{0:R} \in \mathbb{R}^{D_y \times R}$ are used to obtain an appropriate initial hidden state. We consider trainable models $f_{\theta}^\text{h}, C_{\theta}^\text{h}, f_{\nu}^\text{l}, C_{\nu}^\text{l}$ with corresponding rollout mappings $F_{\theta}^\text{h}$ and $F_{\nu}^\text{l}$ (cf. Eq. \eqref{eq:trajectory}), an up/downsampling ratio $k$ and a complementary filter pair $L,H$ (cf. Sec. \ref{sec:complementary}). The weights $\theta$ and $\nu$ are trained by minimizing the root-mean-squared error (RMSE) $\Vert y-\hat{y} \Vert_2$ via 
\begin{equation} \label{eq:rec_training}
\begin{aligned}
\hat{\theta} & = \arg\min_{\theta} \Vert H(\hat{y})_{R:N} - F^\text{h}_{\theta}(\hat{y}_R^\text{h},h_R^\text{h},N-R) \Vert_2\\
\hat{\nu} & = \arg\min_{\nu} \Vert d_k(L(\hat{y})_{R:N})-F^\text{l}_{\nu}(\hat{y}_R^\text{l},h_R^\text{l},\tilde{N})\Vert_2,
\end{aligned}
\end{equation}
with $\hat{y}_R^\text{h}=H(\hat{y})_R$, $\hat{y}_R^\text{l}=L(\hat{y})_R$, $N-R$ steps $H(\hat{y})_{R:N}$ and $L(\hat{y})_{R:N}$ from the filtered signals $H(\hat{y})$ and $L(\hat{y})$ and $\tilde{N}=\lfloor (N^{\prime}-R)/k \rfloor$.
The hidden states $h_R^\text{h}$ and $h_R^\text{l}$ are obtained from a warmup phase that we specify in the appendix. 
 
\paragraph{Predictions:  }
A prediction with $N^{\prime}-R$ steps $\tilde y_{R:N^{\prime}}$ is obtained by adding the high-frequency predictions and the upsampled low-frequency predictions 
\begin{equation} \label{eq:rec_predictions}
\tilde{y}_{R:N^{\prime}}=F^\text{h}_{\theta}(\tilde{y}_R^\text{h},h_R^\text{h},N^{\prime}-R)+u_k(F_{\nu}^\text{l}(\tilde{y}_R^\text{l},h_R^\text{l},\tilde{N})),
\end{equation}
where $\tilde{y}_R^\text{h}=H(\tilde{y}_{0:R})$ and $\tilde{y}_R^\text{l}=L(\tilde{y}_{0:R})$ have to be provided and $\tilde{N}=\lfloor (N^{\prime}-R)/k \rfloor$. The hidden states $h_R^\text{h}$ and $h_R^\text{l}$ can, for example, be obtained from a short warmup phase. 

A slight modification of the method is obtained by wrapping an additional high-pass filter around the predictions via $H(F^\text{h}_{\theta}(\tilde{y}_R^\text{h},h_R^\text{h},N-R))$ in Eq. \eqref{eq:rec_training} during training and via $H(F^\text{h}_{\theta}(\tilde{y}_R^\text{h},h_R^\text{h},N^{\prime}-R))$ in Eq. \eqref{eq:rec_predictions} during predictions.
This adds an additional guarantee preventing the high-frequency model from producing low-frequency errors. 
We provide a numerical comparison of both variants in our experiments.
\subsection{Hybrid modeling}
In the hybrid case, we assume access to a simulator that produces predictions $y^\text{s}_{0:N}$. 
Thus, reliable long-term predictions are already available. 
In this case, we can directly train a single recurrent model in an end-to-end fashion (see Figure \ref{fig:scheme} (right)).
In particular, the low-pass filter is applied to the simulator, whereas the high-pass filter is applied to the learning-based trajectory. 
Training directly with the complementary filter ensures that each model indeed stays on its time scale. 
By decoupling the propagation of latent states and the filtered simulator states, the method is technically applicable to a large class of simulators. 
It is solely required that the simulator is able to produce time-series predictions of the system given initial conditions. 
Differentiating through the simulator or any insight into the simulator's hidden states is not required.

\paragraph{Training and predictions: }
Consider training data $\hat{y}_{0:N} \in \mathbb{R}^{D_y \times N}$, a trainable model $f_{\theta}, C_{\theta}$ with corresponding rollout mapping $F_{\theta}$ (cf. Eq. \eqref{eq:trajectory}) and a complementary filter pair $(L,H)$. Again, the first $R$ steps $\hat y_{0:R}$ are used for providing the intial hidden state $h_R$. The weights $\theta$ are trained via 
\begin{equation}\label{eq:loss}
\hat{\theta} = \arg \min_{\theta} \Vert H(y^{\text{r}}) + L(y^\text{s}_{R:N}) - \hat{y}_{R:N} \Vert_2,
\end{equation} 
with $y^{\text{r}}=F_{\theta}(\hat{y}_R,h_R,N-R)$. The calculation of $H(y^{\text{r}}) + L(y^\text{s}_{R:N})$ can directly be obtained by Eq. \eqref{eq:IIR}.

\subsection{Filter design}
We design filters $H$ and $L$ (cf. Eq. \eqref{eq:rec_training} and \eqref{eq:loss}) before training. 
In the purely learning-based scenario, a broad range of cutoff frequencies is possible, which we demonstrate empirically in the appendix.
In the hybrid case, we aim to use as much correct long-term information as possible from the simulator without including short term errors.  
In general, suitable cutoff frequencies can often be derived from domain knowledge.
Here, we analyze the frequency spectra of ground truth and simulator in order to find a suitable cutoff frequency. 
For a specific filter design, we test the plausibility of the complementary filter by applying the high-pass component to the measurements and the low-pass component to the simulator.  
Calculating the RMSE between the combined signal and ground truth indicates whether the filters are appropriate. 
For a more detailed introduction to general filter design, we refer the reader to \citet{oppenheim1999discrete}.

\begin{figure*}[tb]
\centering
\includegraphics[width= \textwidth]{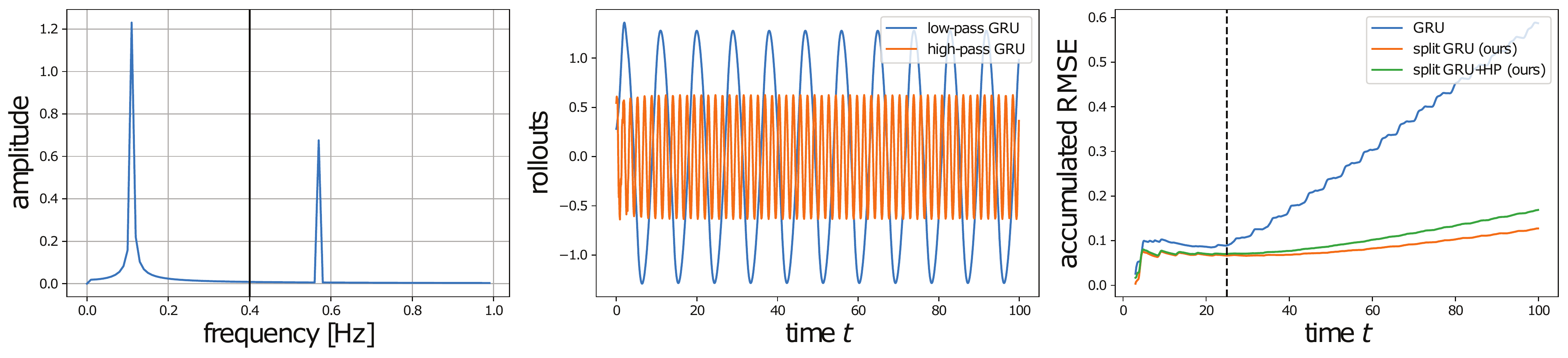}
	\caption{Demonstrating the method with the double-mass spring system (i). Shown is the analysis of the frequency spectrum with marked cutoff frequency (left). This yields the predictions of the two seperate GRUs (middle).
	The accumulated RMSE over time indicates good short-and-long term behavior, while the baseline method accumulates errors (right).}
\label{fig:mass}
\end{figure*}

\begin{figure*}[tb]
	\centering
	\includegraphics[width= \textwidth]{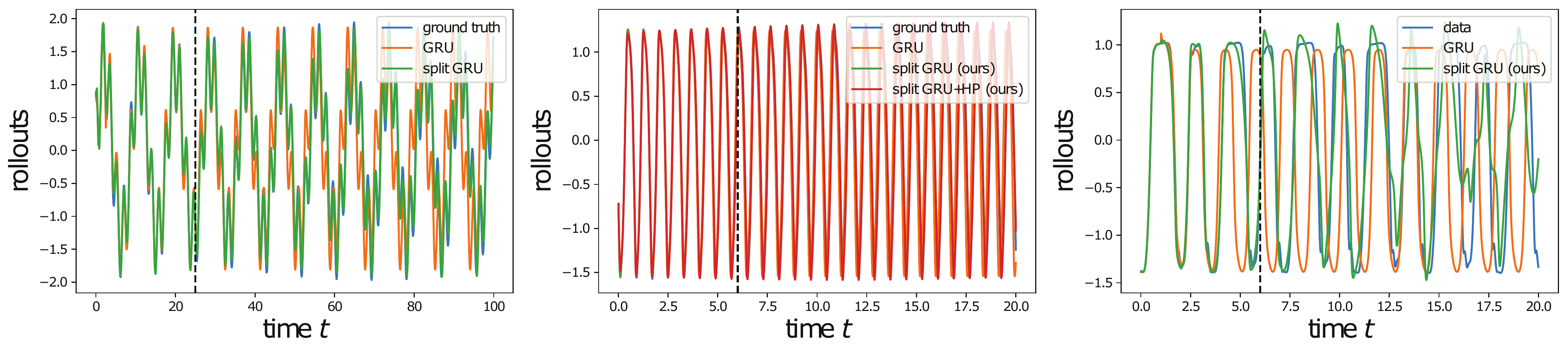}
	\caption{Rollouts for the purely learning-based scenario with all three methods for Systems (i) (left), (ii) (middle) and (iv) (right). The training horizon is marked with dotted lines. The results show accumulating errors of the baseline method in contrast to our approach.}
	\label{fig:learning_based}
\end{figure*}

\begin{figure*}[tb]
	\centering
	\includegraphics[width= \textwidth]{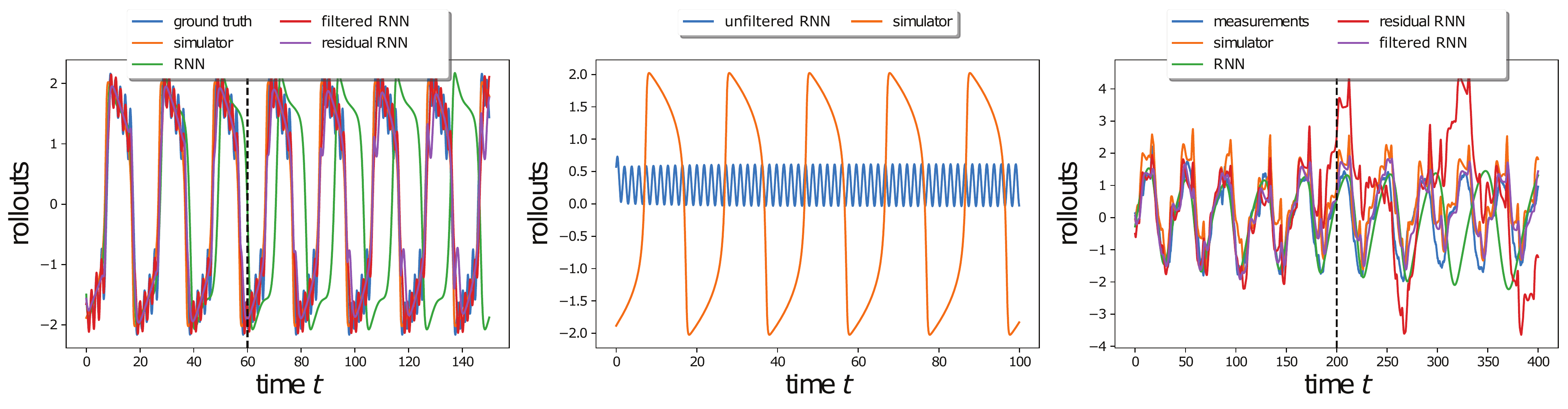}
 	\caption{Rollouts for the hybrid setting. Shown are the results for the Van-der-Pol oscillator (v) with RNN (left) and rollouts of the single components before combining them via the complementary filter (middle). Further, the rollouts of the drill-string system (vi) with RNN are shown (right). The training horizon is marked with dotted lines. The results demonstrate accumulating errors in the baseline methods, while our approach provides accurate short and long-term predictions.}
	\label{fig:hybrid}
\end{figure*}

\section{Experiments} \label{section:exp}
In this section, we demonstrate that our complementary filter-based methods yield accurate long and short-term predictions on simulated and real world data. 
In the hybrid setting, we consider additional access to a physics-based simulation that is able to predict the long-term behavior of the system but is not capable of accommodating all short-term details due to e.g., modeling simplifications.

\subsection{Baselines}
We consider four systems.
For each system, we have access to measurement data. 
Either real measurements are available, or we simulate trajectories from the ground truth system and corrupt them with noise.  

We consider the following baselines.

\textbf{RNN: } 
RNN structure that corresponds to an MLP that is propagated through time.

\textbf{GRU: } 
state-of-the-art recurrent architecture for time-series learning \citep{cho-etal-2014-learning}. 

\textbf{Simulator: } 
in the hybrid setting, access to simulator predictions $y^\text{s}$ is required. 

\textbf{Residual GRU/ RNN: }
in the hybrid case, we consider a residual model that combines RNN or GRU predictions $y^\text{r}$ with simulator predictions $y^\text{s}$ via $y=y^\text{r}+y^\text{s}$.

\subsection{Constructing the filters}
We use the tools for IIR filter design provided by \texttt{Scipy} \citep{2020SciPy-NMeth} and apply Butterworth filters.
We construct the coefficients $b_k$ and $a_k$ for the low-pass filter and coefficients $\tilde{b}_k$ and $\tilde{a}_k$ for the high-pass filter as described in Sec. \ref{section:background}, where both filters share the cutoff frequency. 
An example of a frequency spectrum and choice of the cutoff frequency is shown in Figure \ref{fig:mass} (left). 
In the appendix, we add information on the specific design of the complementary filter pairs for each experiment. 
Further, we add frequency spectra for each system.

\subsection{Learning task and comparison}
For each system, we observe a single trajectory. 
The models are trained on a fixed subtrajectory of the full trajectory.
Predictions are performed by computing a rollout of the model over the full trajectory. 
We evaluate the model accuracy by computing the RMSE along the full trajectory. 
On the simulated systems, the RMSE between predictions and ground truth is computed.
On real world data, the RMSE between predictions and measurements is computed. 
Runtimes are reported in the appendix.

\subsection{Purely learning-based model: }
We apply the strategy derived in Sec. \ref{section:method} to GRU models (referred to by “split GRU“) and compare to a single GRU model trained on the entire bandwidth.
In order to draw a fair comparison, we choose an equal number of total hidden units for the baseline GRU and the sum of hidden states in our approach.  
We provide architecture details in the appendix. 
Furthermore, we optionally wrap an additional high-pass filter around the predictions $F^h_{\theta}(\hat{y}_R^h,h_R^h,N-R)$  during training and inference (cf. Eq. \eqref{eq:rec_training} and \eqref{eq:rec_predictions}), and denote this by the suffix "+HP".  
In order to demonstrate the flexibility of our method, we add results with varying cutoff frequencies and downsampling ratios in the appendix.
We train our model on the following systems:

\textbf{(i) Double-mass spring system: } 
We simulate a double-mass spring system that consists of two sinusoidal waves with different frequencies and corrupt the simulation with additional observation noise. 
Training is performed on an interval of 250 steps, while predictions are computed on 1000 steps (further details can be found in the appendix).

\textbf{(ii) - (iv) Double torsion pendulum: }
In the second set of experiments, we consider real measurements from the double-torsion pendulum system introduced in \citet{Lisowski}.
Data are obtained by exciting the system with different inputs. 
In particular, we consider 4 different excitations with varying frequencies.  
Training is performed on the first 600 measurements, while predictions are performed on an 2000-steps interval.

\subsection{Hybrid model}
For the hybrid model, we train our complementary filtering method with GRU and RNN and compare against the corresponding non-hybrid models (GRU and RNN), the corresponding residual models (residual GRU/ RNN), and the simulator. We consider the following systems:

\textbf{(v) Van-der-Pol oscillator: }
Data from a Van-der-Pol oscillator with external force is simulated from the four-dimensional ground truth system (\citet{Cartwright}). 
It is assumed that only the first dimension, corresponding to the position, is observed.
Simulator data are obtained from an unforced Van-der-Pol oscillator.
For the corresponding equations, we refer to the appendix. 

\textbf{(vi) Drill-string: } 
We consider measurement data from the drill string experiment provided in \citet{AARSNES2018712} Figure 14 as training data and the corresponding simulated signal as simulator. 

\subsection{Results}
The results indicate the advantage of leveraging complementary filters for dynamics model learning.
In particular, the resulting predictions show stable short and long-term behavior, while especially the GRU and RNN baselines tend to drift on the long-term due to accumulating errors. 
For both scenarios, we provide additional plots showing the accumulated RMSE over time for each system in the appendix. 

\begin{table*}[!htpb]
	\centering 
	
	\begin{tabular}{rccc}
		\noalign{\smallskip} \hline \hline \noalign{\smallskip}
		System & GRU & split GRU (ours) & split GRU + HP (ours)  \\
		\hline
		(i) & 0.587 (0.002) & \textbf{0.127} (0.008) & 0.168 (0.03)\\
		(ii) & 1.124 (0.485) &  0.331 (0.065) & \textbf{0.318} (0.089) \\
		(iii) & 0.287 (0.15) & 0.159 (0.051)  & \textbf{0.13} (0.02)\\
		(iv)  & 0.262 (0.17) & \textbf{0.201} (0.07) & 0.18 (0.06) \\
		\noalign{\smallskip} \hline \noalign{\smallskip}
	\end{tabular}\quad
	% \flushleft
	\caption{Total RMSEs (mean (std)) over 5 indep. runs with purely learning-based scheme.}
	% \flushleft
	\label{t:errors}	
\end{table*}

\paragraph{Purely learning-based}
The results in Table \ref{t:errors} indicate the advantage of our approach due to accumulating errors for the baseline method. 
Integrating a small model error in each time-step leads to a long-term drift that can also be directly observed in the rollouts (cf. Figure \ref{fig:learning_based}).
Our approach on the other hand does not suffer from this drift due to the specific architecture and therefore outperforms the baseline method on every task.  
The findings are also supported by the RSME over time $(e_n)_{n=0}^N$ with $e_n= \sqrt{\sum_{k=0}^n \frac{1}{n+1} \Vert y_k-\hat{y}_k\Vert^2}$ shown in Figure \ref{fig:mass} (right). 
In some cases our methods yields faster convergence than the baseline method. 
For System i) we report the results after 300 training epochs for our method, while the GRU was trained on 2000 epochs. 
To provide more insights, we demonstrate the functionality of our method with the double-mass spring system (i) (cf. Figure \ref{fig:mass}).
Designing the filters shown in Figure \ref{fig:mass} (left) yields seperate predictions from the two GRUs in Figure \ref{fig:mass} (middle).
Similar results of our split GRU and our split GRU+HP indicate that the most effective part is already contained in the split GRU (cf. Table \ref{t:errors}).
Here, the high-frequency model already stays on the desired time scale and the additional high-pass filter rather introduces a small distortion.
Further, our split GRU+HP shows a higher error in the beginning due to transient behavior of the filter, which can be seen in Figure \ref{fig:mass} (right).
However, the additional high-pass filter guarantees that the high-frequency predictions are indeed affecting the correct time scale.

\begin{table*}[h!]
	\centering 
	\begin{tabular}{rcccc}
		\noalign{\smallskip} \hline \hline  \noalign{\smallskip}
		System & RNN & residual RNN & simulator & filtered RNN (ours) \\
		\hline
		(v) & 1.29 (0.63) & 0.417 (0.03) &0.418&  \textbf{0.347} (0.041) \\
		(vi) & 1.1 (1.26)  & 3.60 (1.62)  & 0.729 & \textbf{0.487} (0.381) \\
		\noalign{\smallskip} \hline \noalign{\smallskip}
	\end{tabular}\quad
	\caption{Total RMSEs for the hybrid model with RNN (mean (std)) over 5 indep. runs.}
	\label{t:MLP}
\end{table*}

\begin{table*}[h!]
	\centering 	
	
	\begin{tabular}{rcccc}
		\noalign{\smallskip} \hline \hline  \noalign{\smallskip}
		System & GRU & residual GRU & simulator & filtered GRU (ours) \\
		\hline
		(v) & 0.463 (0.305)  &  0.476 (0.096)  & 0.418 &  \textbf{0.387} (0.026) \\
		(vi) & 1.140 (0.258)  & \textbf{0.681} (0.055)   & 0.729 &  0.765 (0.008) \\
		\noalign{\smallskip} \hline \noalign{\smallskip}
	\end{tabular}\quad
	\caption{Total RMSEs for the hybrid model with GRU (mean (std)) over 5 indep. runs.}
	\label{t:GRU}
\end{table*}

\paragraph{Hybrid model} 
We report the RMSEs for the hybrid setting with RNNs in Table \ref{t:MLP} and with GRUs in Table \ref{t:GRU}. 
The results demonstrate that our method is beneficial for different types of models, here MLP-based RNNs and GRUs. 
Again, the standard training with single GRU or single RNN shows some drift causing bad long-term behavior.
The unstable long-term behavior is demonstrated particularly clearly by the RNN results shown in Figure \ref{fig:hybrid} (left and right).
While the residual RNN baseline does not suffer from the typical drift that is observed for the RNN baseline, it still shows instabilities in the long-term behavior.
In particular, the results for System (vi) in Figure \ref{fig:hybrid} (right) demonstrate that low-frequency errors occur for the residual model as well. 
Our method, in contrast, eliminates these errors by design. 
However, on System (vi), our filtered GRU is outperformed by the residual GRU since our predictions stay close to the simulator predictions. 
We provide additional insights into our method by depicting the RNN and simulator predictions before combining them via the complementary filter for System (v) in Figure \ref{fig:hybrid} (middle).
Additional plots are provided in the appendix.

\section{Conclusion} \label{section:conclusion}
In this paper, we propose to combine complementary filtering with dynamics model learning. 
In particular, we fuse the predictions of different models, where one models provides reliable long-term predictions and the other reliable short-term predictions. 
Leveraging the concept of complementary filter pairs yields a model that combines the best of both worlds. 
Based on this idea, we propose a purely learning-based model and a hybrid model.
In the hybrid scenario, the long-term predictions are addressed by a simulator, whereas in the purely learning-based scenario an additional model has to be trained.  
The experimental results demonstrate that our approach yields predictions with accurate long and short-term behavior. 
An interesting topic for future research is an extension of the hybrid scenario learning the relationship between simulator predictions and learning-based predictions.

\section*{Acknowledgements} \label{sec:acknowledgements} 
The authors thank Karim Barsim, Alexander Gr\"afe and Andreas Ren\'{e} Geist for valuable discussions and feedback. 
Furthermore, we thank Pawe\l \,Olejnik for providing measurement data from the double-torsion pendulum for the systems (ii)-(iv) that we consider in our experiments.   
%\FloatBarrier
%\clearpage
\bibliography{bib} 

\end{document}

\maketitle

% !TEX root = neurips_2021.tex
\section{Background} \label{section:filter_background}
In this section, we provide the derivation for the filtering theory used in Sec. 2 and Sec. 3. 
In particular, we derive the recurrence equation from the corresponding transfer functions in the z-domain for the IIR filter and the complementary filter. 
For a more detailed introduction to digital filter design, we refer the reader to \citet{oppenheim1999discrete}.
\subsection{Z-transform}
The z-transform \citep{ztransform} can be interpreted as discrete-time version of the Laplace transform.
It is designed in order to analyze and manipulate discrete signals in the frequency domain. 
Let $(y_n)_{n=0}^N$ be a discrete-time signal. For the z-transform $\mathcal{Z}(y): \mathbb{C} \to \mathbb{C}$ of $y$ it holds that 
\begin{equation}
\begin{aligned}
\mathcal{Z}(y)(z) = Y(z) = \sum_{n=1}^{\infty} y_n z^{-n},
\end{aligned}
\end{equation} 
where $z$ represents the frequency components and their corresponding amplitudes. Similar to the Fourier transform and Laplace transform, the z-transform has useful properties that are leveraged for filter design. 
We specifically leverage the time delay property 
\begin{equation}\label{eq:time_delay}
\mathcal{Z}(y_{n-k})=z^{-k}Y(z).
\end{equation}
Here we use $y_{n-k}$ as a notation for the original signal $y$ shifted $k$ steps.
\subsection{Filter design}
Theoretically, a filter is obtained by designing a transfer function $F:l^{\infty}\rightarrow l^{\infty}$ in the frequency domain. 
In order to obtain the representation in the time domain, the following steps are performed:
\begin{itemize}
\item design a filter transfer function $\mathcal{F}$ in z-domain according to desired properties.
\item apply z-transform to signal $Y = \mathcal{Z}(y)$. 
\item multiply filter transfer function to z-transformed signal $\tilde{Y} = \mathcal{F} Y$.
\item apply inverse z-transform to obtain filtered signal in time domain $\tilde{y} = \mathcal{Z}^{-1}(\tilde{Y})$. 
\end{itemize}
\subsection{IIR-filter} 
In this section, we derive the recurrence equation for the IIR filter $H$ with transfer function $\mathcal{H}$ given in the form
\begin{equation}\label{eq:coeff}
\mathcal{H}(z)=\frac{\sum_{k=0}^P b_k z^{-k}}{\sum_{k=0}^Q a_k z^{-k}}.
\end{equation}
Here, we will consider filters with $P=Q$. Clearly, such a formulation can be derived without loss of generality by adding as many zero coefficients as nescessary.   
The coefficients in Eq. \eqref{eq:coeff} are obtained from the desired filter properties in the frequency spectrum. 
There are different strategies for constructing the parameters with certain advantages and disadvantages. 
By applying the above strategy to the signal $y$, a recurrence equation for the filtered signal $\tilde{y}$ can be obtained. 
Multiplying the transfer function $\mathcal{H}(z)$ with the z-transformed signal $Y(z)$ results in the filtered signal in the frequency domain  
\begin{equation}\label{eq:IIR_der}
\begin{aligned}
\tilde{Y}(z)=\frac{\sum_{k=0}^P b_k z^{-k}}{\sum_{k=0}^P a_k z^{-k}} Y(z).
\end{aligned}
\end{equation}
In order to derive the corresponding recurrence equation in the time-domain, Eq. \eqref{eq:IIR_der} is multiplied with the denominator resulting in 
\begin{equation}\label{eq:mult}
\left(\sum_{k=0}^P a_k z^{-k}\right)\tilde{Y}(z)=\left(\sum_{k=0}^P b_k z^{-k}\right)Y(z).
\end{equation}
Transforming back to the time domain yields 
\begin{equation}\label{eq:inv}
\mathcal{Z}^{-1}\left(\sum_{k=0}^P a_k z^{-k}\tilde{Y}(z)\right)=\mathcal{Z}^{-1}\left(\sum_{k=0}^P b_k z^{-k} Y(z)\right).
\end{equation}
Applying the time-delay rule \eqref{eq:time_delay} and the linearity of the z-transform yields 
\begin{equation}\label{eq:time}
\sum_{k=0}^P a_k \tilde{y}_{n-k}=\sum_{k=0}^P b_k y_{n-k}.
\end{equation}
Solving for $\tilde y$ yields a recurrence equation for the signal 
\begin{equation}\label{eq:rearange}
\tilde y_n=\frac{1}{a_0}\left(\sum_{k=1}^P a_k \tilde{y}_{n-k}+ \sum_{k=0}^P b_k y_{n-k}\right),
\end{equation}
where $P$ describes the filter order. 
Eq. \eqref{eq:rearange} can be applied for $n\geq P$. For $n\leq P$ an initialization technique has to applied. 
The filter can, for example, be initialized with zeros or the original signal.   

\paragraph{Forward-backward filter: }
IIR-filters suffer from a phase delay yielding a slightly shifted signal. 
This effect can be eliminated by applying the identical IIR-filter twice. Once in forward- and once in backward direction. I.e., the final result is obtained by reversing the filtered signal $\tilde y$, filtering this again via Eq. \eqref{eq:rearange} and doing another reverse. 
Technically, this results in an IIR-filter with higher order and specific initialization strategy \citep{492552}. 

\subsection{Complementary filter} 
In this section, we derive the equation for the complementary filter.
Given are the two filters $H$ and $L$ with IIR transfer functions $\mathcal{H}$ and $\mathcal{L}$. 
Consider a high-pass filter $\mathcal{H}$ with coefficients $a_k,b_k$ and a low-pass filter $\mathcal{L}$ with coefficients $a_k, \tilde{b}_k, k=1,\dots P$.
For notational simplicity we chose the same denominator and the same order $P$ for both filters. This can be done without loss of generality.
Applying the transfer functions to the two transformed signals $Y^{\text{h}}$ and $Y^{\text{l}}$ yields   
\begin{equation}
\begin{aligned}
\tilde{Y}(z) & = \mathcal{H}(z) Y^{\text{h}}(z) + \mathcal{L}(z)Y^{\text{l}}(z) \\
&= \frac{\sum_{k=0}^P b_k z^{-k}}{\sum_{k=0}^P a_k z^{-k}} Y^{\text{h}}(z)+ \frac{\sum_{k=0}^{P}  \tilde{b}_k z^{-k}}{\sum_{k=0}^P a_k z^{-k}}Y^{\text{l}}(z).    
\end{aligned}
\end{equation} 
Multiplying with the denominator as in Eq. \eqref{eq:mult} yields 
\begin{equation}
\begin{aligned}
\tilde{Y}(z) \sum_{k=0}^P a_k z^{-k} = Y^{\text{h}}(z) \sum_{k=0}^P b_k z^{-k} + Y^{\text{l}}(z) \sum_{k=0}^{\tilde{P}} \tilde{b}_k z^{-k}
\end{aligned}
\end{equation}
Similar to Eq. \eqref{eq:inv}, we apply the time delay rule \eqref{eq:time_delay} and the inverse z-transform which yields
\begin{equation} \label{eq:IIR}
\sum_{k=0}^P a_k \tilde{y}_{n-k} =  \left(\sum_{k=0}^P b_k y^{\text{h}}_{n-k}+\sum_{k=0}^{P}  \tilde{b}_k y^{\text{l}}_{n-k}\right).  
\end{equation}
Solving for $\tilde y_n$ yields
\begin{equation}\label{eq:rearangeComp}
\tilde y_n=\frac{1}{a_0}\left(\sum_{k=1}^P a_k \tilde{y}_{n-k}+ \sum_{k=0}^P b_k y^{\text{h}}_{n-k}+\sum_{k=0}^{P}  \tilde{b}_k y^{\text{l}}_{n-k}\right).
\end{equation}
Again, Eq. \eqref{eq:rearangeComp} can be applied for $n\geq P$. For $n\leq P$ an initialization technique has to applied. 
The filter can, for example, be initialized with zeros or one of the input signals. 
In our case, the filter is initialized via the training signal. 

\paragraph{Perfect complement} \label{section:perfect}
To construct perfectly complementary filter pair $H$, $L$ for a given transfer function $\mathcal{H}$ with coefficients $a_k,b_k$, we can chose $\tilde{a}_k=a_k$ and $\tilde{b}_k=a_k-b_k$ as coefficients for $\mathcal{L}$. In this case it holds that $y=H(y)+L(y)$ since
\begin{equation}\label{eq:identity}
\begin{aligned}
\tilde{Y}(z)&= \frac{\sum_{k=0}^P b_k z^{-k}}{\sum_{k=0}^P a_k z^{-k}}Y(z)+\frac{\sum_{k=0}^P (a_k-b_k) z^{-k}}{\sum_{k=0}^P a_k z^{-k}}Y(z)=Y(z).
\end{aligned}
\end{equation}

\subsection{Nyquist frequency} 
In the purely learning-based scenario, we train on the downsampled signal and apply an upsampling technique for predictions. 
Low-pass filtering the training signal ensures that downsampling causes no loss of information. 
Admissable sampling rates $f_{\textrm{sample}}$ can be obtained via the Nyquist-Shannon theorem (see \citet{Shannon1949}).
In general for the cutoff frequency $f_w$ it has to hold 
\begin{equation}\label{eq:nyquist}
f_w < f_{\textrm{nyquist}},
\end{equation}
with $f_{\textrm{nyquist}}=\frac{1}{2} f_{\textrm{sample}}$.
Thus, it has to be assured that the sampling rate after downsampling fulfills Eq. \eqref{eq:nyquist}.
\subsection{Recurrent dynamics model learning}
We specify the learning details for the recurrent dynamics model learning in Sec. 4.1. 
In particular, we demonstrate the transition function for a GRU and an MLP and add details on obtaining the initial hidden state. 

\paragraph{MLP training: }
An MLP with dynamics $f_{\theta}$ and linear observation model $C_{\theta}$ is trained via Euler steps on the hidden layers
\begin{equation}\label{eq:RNN}
\begin{aligned}
h_{n+1} & = h_n+\Delta_t f_{\theta}(h_n) \\
y_n & = C_{\theta} h_n,
\end{aligned}
\end{equation}
with initial state $h_0$ and step size $\Delta_t$.
For the MLP, we train a recognition model $r_{\theta}$ that is inspired by \citet{pmlr-v80-doerr18a}.
The recognition model consists of an additional MLP that estimates the latent state $h_R$ from the first $R$ noisy observations $\hat{y}_{0:R}$ via 
\begin{equation}
h_R = r_{\theta}(\hat{y}_{0:R})
\end{equation} 

\paragraph{GRU training: }
The GRU with transition dynamics $f_{\theta}$ is trained via 
\begin{equation}\label{eq:GRU}
\begin{aligned}
h_{n+1} & = f_{\theta}(h_n,y_n) \\
y_n & =  C_{\theta} h_n,
\end{aligned}
\end{equation} 
The warmup phase over $R$ steps is performed by feedbacking the observations $\hat{y}_n$ instead of the GRU outputs $y_n$ via $h_{n+1} = f_{\theta}(h_n,\hat{y}_n)$.
This yields an appropriate hidden state $h_R$. 

% !TEX root = neurips_2021.tex
\section{Experiment setup} \label{section:experiment_setup}
In this section, we provide the experimental details of our methods. 
In particular, we specify the hyperparameters and filter details. 
The experiments were conducted on a GPU cluster.

\subsection{Purely learning-based method}
We supply algorithmic details for the purely learning-based model. 
\paragraph{Filter design}
The filters coefficients are designed before training. 
We use the tools for IIR filter design provided by Scipy \citep{2020SciPy-NMeth} and apply Butterworth filters with specified cutoff frequency $\texttt{W\_n}$.
Precisely, we use the scipy function \texttt{scipy.signal.iirfilter}. 
First, the frequency spectra are analyzed in order to find a suitable cutoff frequency. 
The cutoff frequency is then used to obtain the complementary filter pair, where high-pass and low-pass filter share the cutoff frequency.
For each experiment, expect the double mass-spring system (i), we consider the filter order $N=1$, which directly yields perfectly complementary filter pairs via the default parameters. 
For System (i) we observed better reconstruction properties of the original training signal by choosing a higher filter order and a non-perfectly complementary filter pair.
The filters are applied by calling the function $\texttt{filtfilt}$ from $\texttt{torchaudio.functional.filtering}$ with the predefined parameters. 
This function applies forward-backward filtering to the signals.  
The exact filter parameters are shown in Table \ref{t:LB}.

\paragraph{GRU training}
For all experiments, we construct a GRU via the pytorch function $\texttt{torch.nn.GRU}$ with default parameters. 
The observation matrix $C_{\theta}$ is chosen as linear layer.
We set $\texttt{input\_size}=1$ and specify the $\texttt{hidden\_size}$ for each experiment. 
%The highpass filter parameters are obtained by calling $\texttt{scipy.signal.iirfilter(1,Wn=w,ftype='butter',btype='lowpass',fs=f)$,
%The corresponding lowpass filter parameters are obtained from the perfect complements.  
%where \texttt{w} corresponds to the cutoff frequency and \texttt{f} to the sampling frequency and is specified below. 
For all experiments we choose Adam optimizer with learning rate $10^{-3}$.
In order to accelerate the training process and prevent the method from overfitting, we split the training trajectory into subtrajectories. 
For all experiments we use batch sizes of $50$. 
Downsampling with rate $k$ is obtained by considering every $k^{th}$ step of the signal. 
Upsampling is obtained via calling the function \texttt{upsample(scale\_factor = k, mode = 'linear', align\_corners=False)} from $\texttt{torchaudio.functional.filtering}$.
This function provides upsampling without adding high-frequency artefacts. 
The exact hyperparameters can be obtained from Table \ref{t:LB}.
\begin{table*}[ht]
	\caption{Hyperparameters for learning-based method}
	\centering
	\label{t:LB}
	\begin{tabular}{rcccc}
		\noalign{\smallskip} \hline \hline    \noalign{\smallskip}
		parameter & i) & ii) & iii) & iv)  \\
		\hline
		\texttt{hidden\_size} (GRU 1)  & 48 & 64 &  64 & 64 \\
		\texttt{hidden\_size} (GRU 2) & 16  & 32 & 32 & 32\\
		cutoff frequency w & 0.4  & 0.07 & 0.1 & 0.2 \\
		filter order & 3 & 1 & 1 & 1 \\
		sample frequency f & 10 & 10 & 10 & 10\\
		subtrajectory length & 150 & 250 & 250 & 250 \\
		warmup phase & 30 & 100 & 50 & 50 \\
		training steps & 300 & 1000 & 501 & 1000 \\
		sampling rate k &2 & 2 & 2 & 2 \\
		\noalign{\smallskip} \hline \noalign{\smallskip}
	\end{tabular}\quad
\end{table*}

\subsection{Hybrid model}
We provide algorithmic details for the hybrid model. 
\paragraph{Filter design: }
The filters are designed before training. 
We use the tools for IIR filter design provided by Scipy \citep{2020SciPy-NMeth} and apply Butterworth filters.
First, the frequency spectra from training data and simulator are analyzed in order to find a suitable cutoff frequency. 
Here we chose $P=1$ for the filter order. 
We construct the coefficients $b_k$ and $a_k$ for a low-pass filter as described and compute the complementary high-pass filter coefficients $\tilde{b}_k =a_k-b_k$ and $\tilde{a}_k=a_k$. 
The low filter order worked well in our experiments and showed good training results. 
For all experiments, we initialize the first $P$ steps of the filtered signal with the first $P$ steps of the training signal.  

\paragraph{MLP training: } 
Our MLP dynamics $f_{\theta}$ (cf. Eq. \eqref{eq:RNN}) consist of input layer, hidden layer and output layer. 
The observation matrix $C_{\theta}$ is chosen $(1,0,\dots,0)$, such that it extracts the first value of the hidden state.  
This can be done without loss of generality. 
We apply $\texttt{tanh}$ activation functions. 
Thus, we end up with an input layer $\texttt{tanh(Lin(input\_dim,hidden\_dim)}$, hidden layer $\texttt{tan(\texttt{Lin}(hidden\_dim,hidden\_dim)}$ and output layer  $\texttt{Lin(hidden\_dim,output\_dim)}$.
For all experiments, we choose the hidden dimension $\texttt{hidden\_dim}=500$.

For the MLP setup, we train a recognition MLP that takes the first $R$ steps of the trajectory as an input and provides the initial hidden state $h_R$ as an output. The recognition MLP consists of an input layer $\texttt{tanh(Lin(n,rec\_dim))}$, a hidden layer $\texttt{tanh(Lin(rec\_dim,rec\_dim))}$ and an output layer $\texttt{tanh(Lin(rec\_dim,input\_dim))}$.
For all experiments we choose the hidden dimension of the recognition model $\texttt{rec\_dim}=100$.
For all experiments, we choose Adam optimizer. 
The exact hyperparameters can be obtained from Table \ref{t:MLP}.

For task iii) (Van-der-Pol oscillator) we choose an initial learning rate $10^{-3}$. 
For our method and the single MLP baseline, the learning rate is reduced to $10^{-4}$ after 20 steps and to $10^{-5}$ after 500 steps. 

For task iv) (Drill-string system) we choose an initial learning rate $10^{-3}$. 
For our method and the basic hybrid model the learning rate is reduced to $10^{-4}$ after 10 steps.

\begin{table*}[ht]
	\caption{Hyperparameters for hybrid model with MLP}
	\centering
	\label{t:MLP}
	\begin{tabular}{rcc}
		\noalign{\smallskip} \hline \hline   \noalign{\smallskip}
		parameter & i) & ii) \\
		\hline
		cutoff frequency w & 0.25   & 0.1 \\
		sample frequency f & 10 & 10 \\
		subtrajectory length & 200 & 1500 \\
		recognition steps $n$ & 10 & 50 \\
		training steps & 2000 & 300\\
		$\texttt{input\_dim}$ & 4 & 15 \\
		\noalign{\smallskip} \hline \noalign{\smallskip}
	\end{tabular}\quad
\end{table*}

\paragraph{GRU training: }
For all experiments, we construct a GRU via the pytorch function $\texttt{torch.nn.GRU}$ with default parameters. 
We set $\texttt{input\_size}=1$ and specify the $\texttt{hidden\_size}$ for each experiment. 
The exact hyperparameters can be obtained from Table \ref{t:GRU}
For all experiments, we choose Adam optimizer with learning rate $10^{-3}$. 

\begin{table*}[ht]
	\caption{Hyperparameters for hybrid model with GRU}
	\centering
	\label{t:GRU}
	\begin{tabular}{rcc}
		\noalign{\smallskip} \hline \hline   \noalign{\smallskip}
		parameter & i) & ii) \\
		\hline
		\texttt{hidden\_size} GRU  & 64 & 64 \\
		cutoff frequency w & 0.25  & 0.1 \\
		sample frequency f & 10 & 10 \\
		subtrajectory length & 200 & 500 \\
		warmup phase & 10 & 50 \\
		training steps & 2000 & 500  \\
		\noalign{\smallskip} \hline \noalign{\smallskip}
	\end{tabular}\quad
\end{table*}

\paragraph{Runtimes: }
We provide results for the runtimes (in seconds) for all experiments in Table \ref{t:errors}. 
For our purely learning-based scheme the runtimes could be further improved by parallelizing the training of the two networks. 
\begin{table*}[ht]
	\centering 
	\caption{Runtimes for purely learning-based scheme}
	% \flushleft
	\label{t:errors}
	\begin{tabular}{rccc}
		\noalign{\smallskip} \hline \hline \noalign{\smallskip}
		task & GRU & split GRU (ours) & split GRU + HP (ours) \\
		\hline
		(i) & 535 & 233 & 234\\
		(ii) & 1017 & 1436 & 1439\\
		(iii) & 497 & 703  & 699\\
		(iv)  & 496 & 702 &705\\
		\noalign{\smallskip} \hline \noalign{\smallskip}
	\end{tabular}\quad
	% \flushleft
	
\end{table*}

\begin{table*}[ht]
	\centering 
	\caption{Runtimes for hybrid model with MLP}
	\label{t:MLP}
	\begin{tabular}{rcccc}
		\noalign{\smallskip} \hline \hline \hline \noalign{\smallskip}
		task & MLP & Residuum MLP & simulator & filtered MLP (ours) \\
		\hline
		(iii) & 1414 & 963 & - &  1867 \\
		(iv) &2161  & 2913 & - & 4283 \\
		\noalign{\smallskip} \hline \noalign{\smallskip}
	\end{tabular}\quad
\end{table*}

\begin{table*}[ht]
	\centering 	
	\caption{Runtimes for hybrid model with GRU}
	\label{t:GRU}
	\begin{tabular}{rcccc}
		\noalign{\smallskip} \hline \hline \hline \noalign{\smallskip}
		task & GRU & Residuum GRU & simulator & filtered GRU (ours) \\
		\hline
		(iii) & 2071  &  2263  & - &  5105 \\
		(iv) & 3831 & 3781   & - &  6797 \\
		\noalign{\smallskip} \hline \noalign{\smallskip}
	\end{tabular}\quad
\end{table*}
 
\section{Model analysis} \label{section:model}
In this section, we provide details on the generation of our data.
For the simulated data, we specify the chosen parameters. 
Additionally, we provide the Fourier spectra of the models in order to give an insight into the behavior of the models and the choice of cutoff frequencies. 
\begin{figure*}[tb]
	\begin{subfigure}[htb]{0.32\textwidth}
		\centering
		\includegraphics[width= \textwidth]{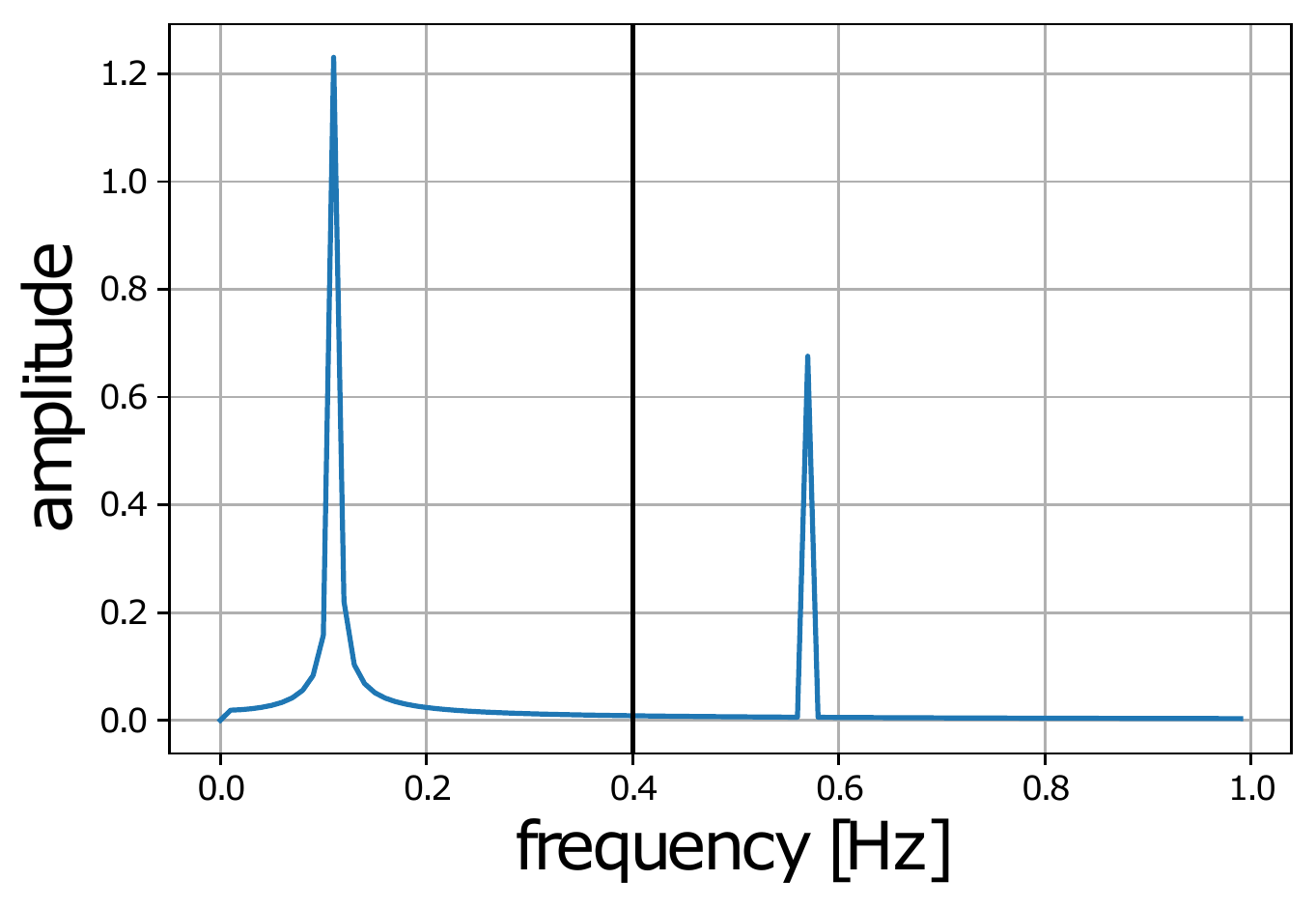}\label{subfig:mass_f}
		\caption{Double-mass spring system (i)}\label{subfig:mass_f}
	\end{subfigure}
	\begin{subfigure}[htb]{0.32\textwidth}
		\centering
		\includegraphics[width= \textwidth]{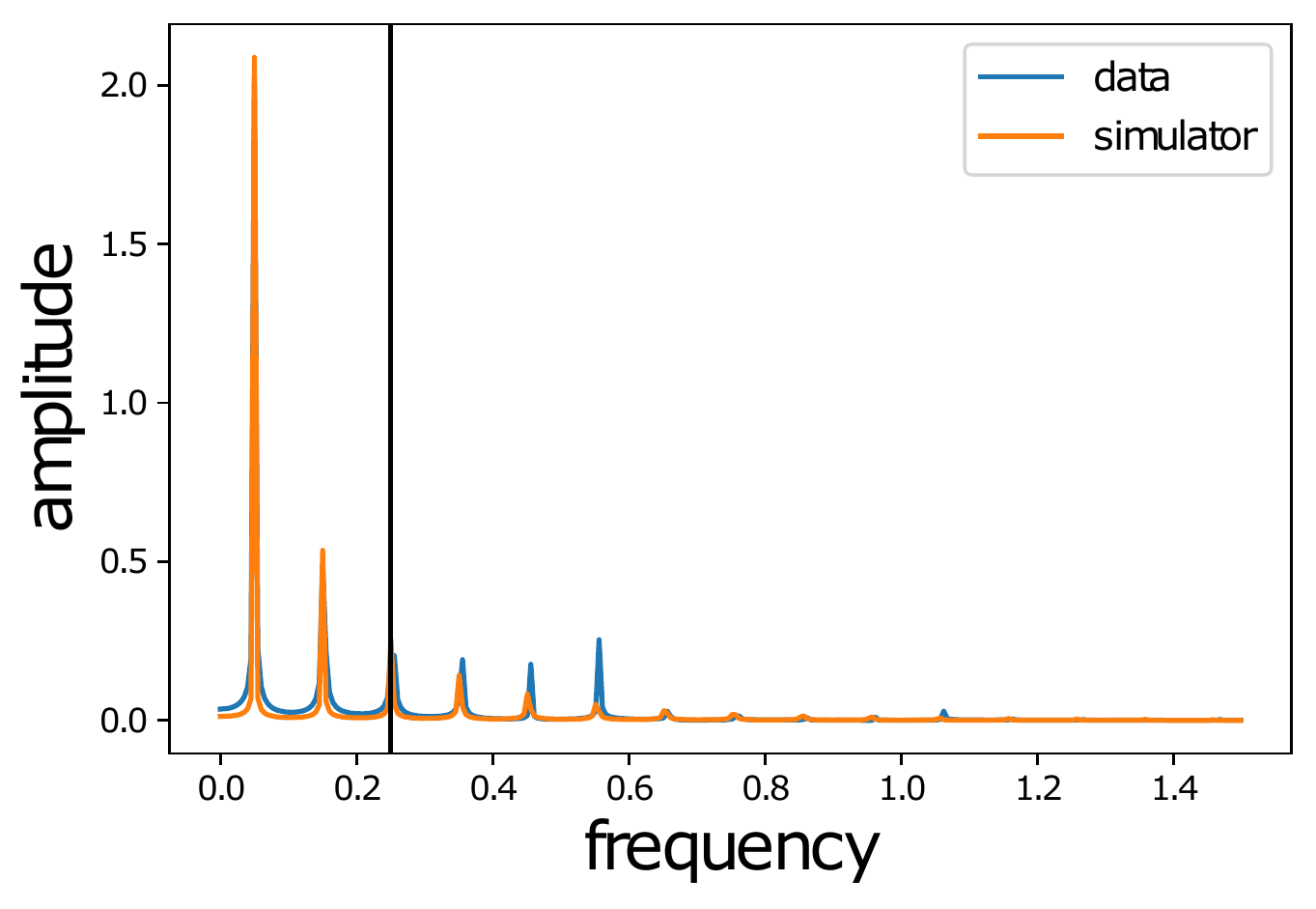}
		\caption{Van der Pol system (v)}\label{subfig:VDP}
	\end{subfigure}	
	\begin{subfigure}[htb]{0.32\textwidth}
		\centering
		\includegraphics[width= \textwidth]{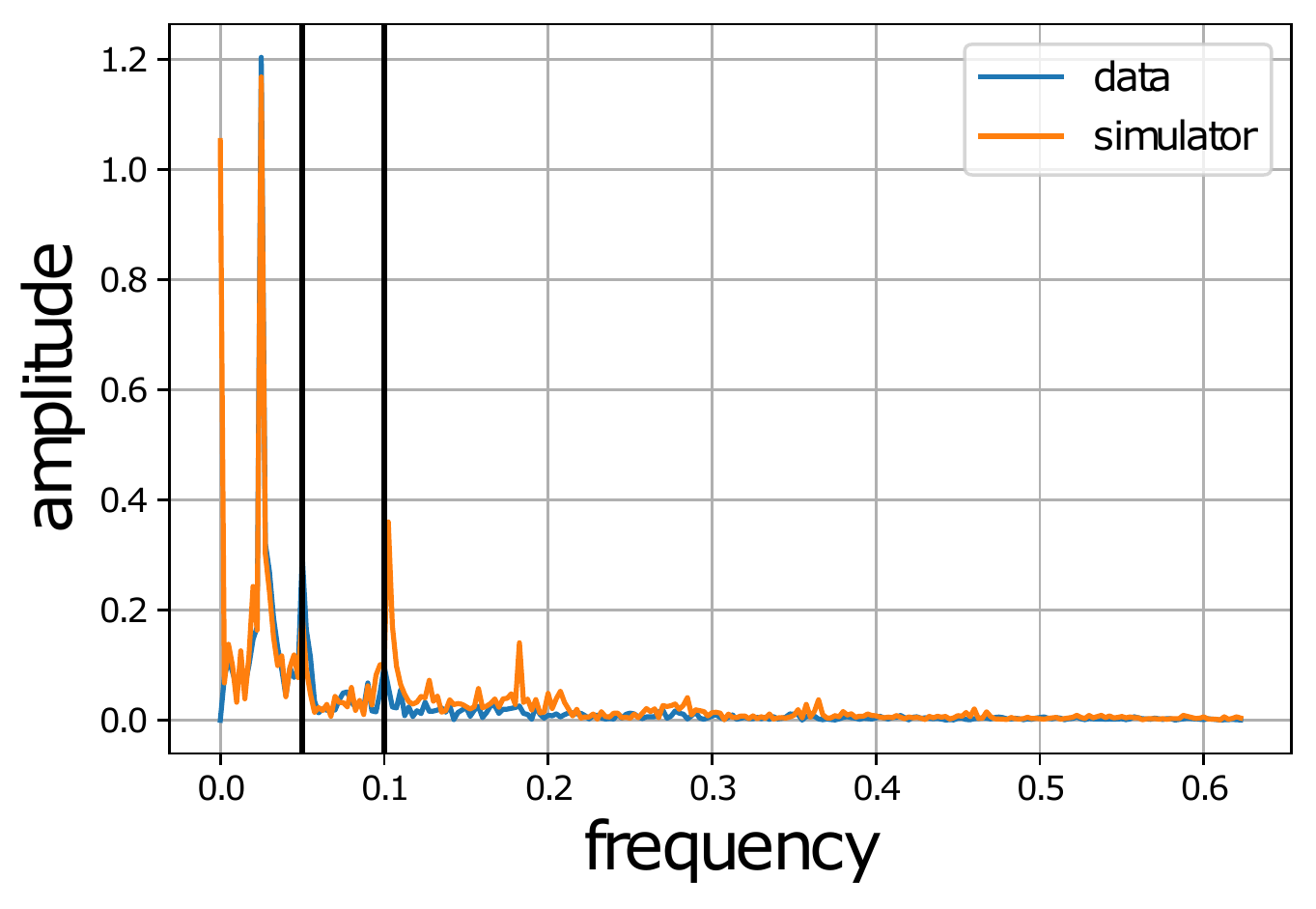}
		\caption{Drill-string system (vi)}\label{subfig:friction}
	\end{subfigure}		
	\caption{Fourier spectra for double-mass spring system in Figure \ref{subfig:mass_f}, Van-der-Pol system in Figure \ref{subfig:VDP} and drill-string system in Figure \ref{subfig:friction}.}
	\label{fig:freq_1}
\end{figure*}

\begin{figure*}[tb]
	\begin{subfigure}[htb]{0.32\textwidth}
		\centering
		\includegraphics[width= \textwidth]{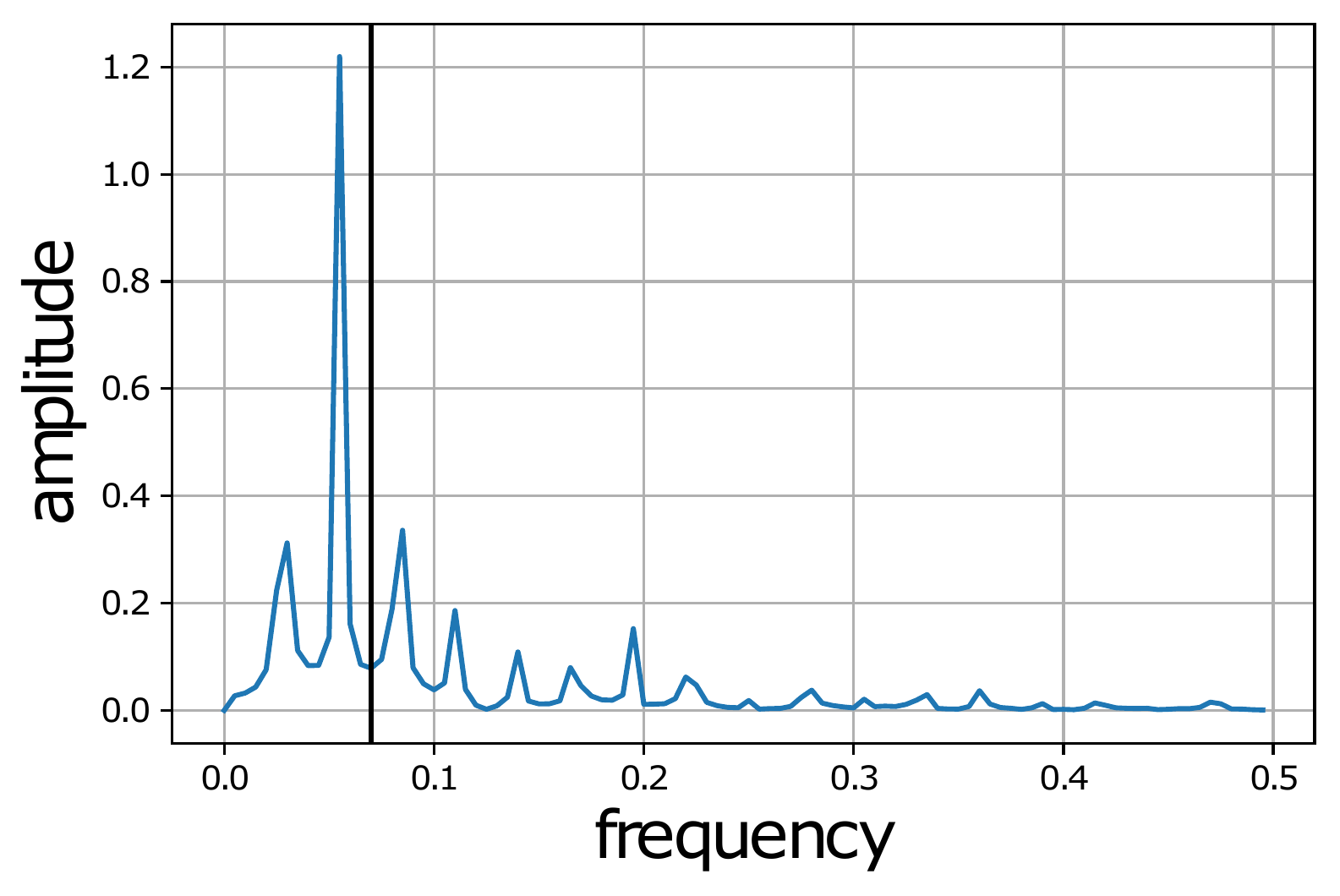}
		\caption{Double-torsion pendulum (ii)}\label{subfig:ii}
	\end{subfigure}
	\begin{subfigure}[htb]{0.32\textwidth}
		\centering
		\includegraphics[width= \textwidth]{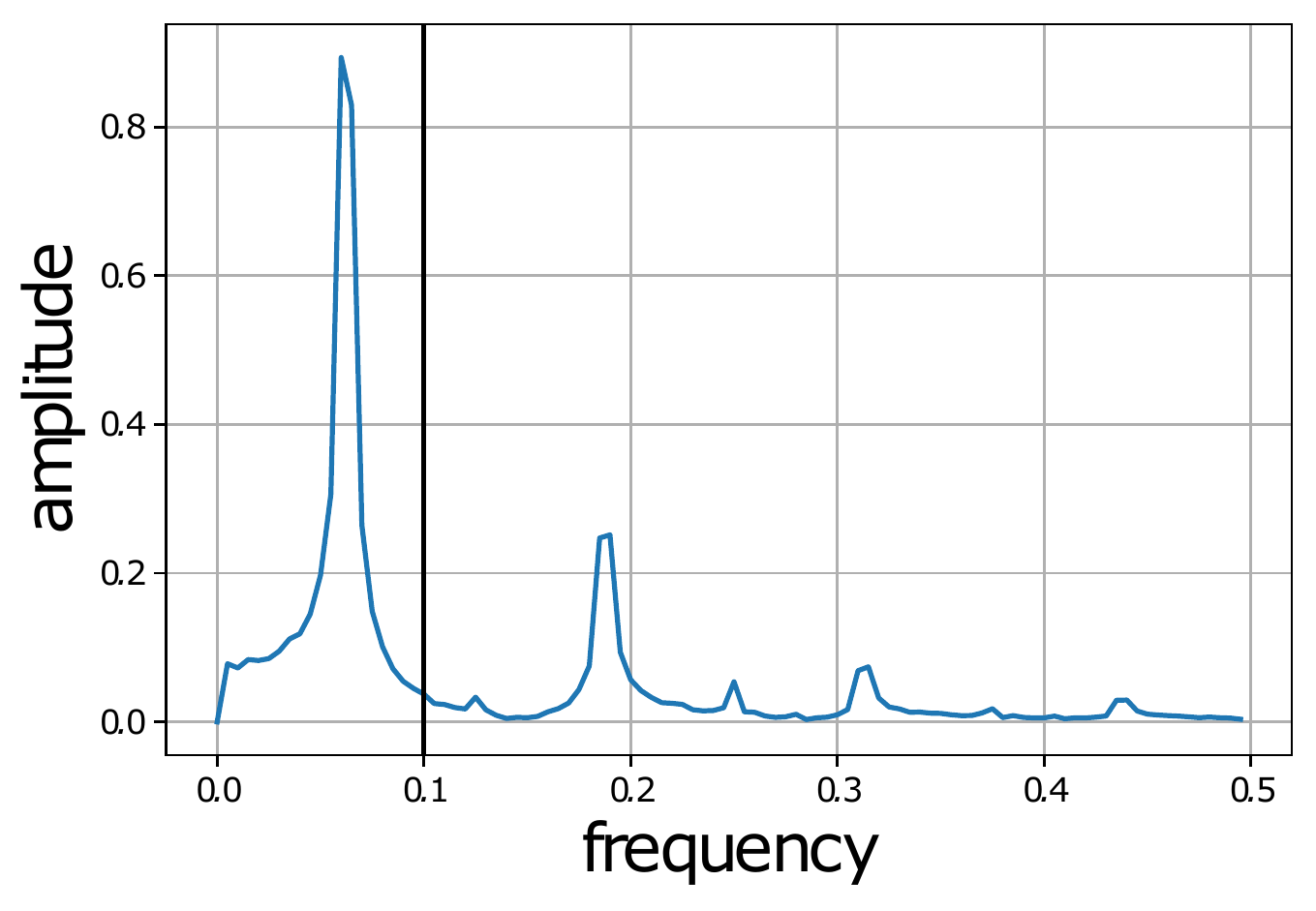}
		\caption{Double-torsion pendulum (iii)}\label{subfig:iii}
	\end{subfigure}	
	\begin{subfigure}[htb]{0.32\textwidth}
		\centering
		\includegraphics[width= \textwidth]{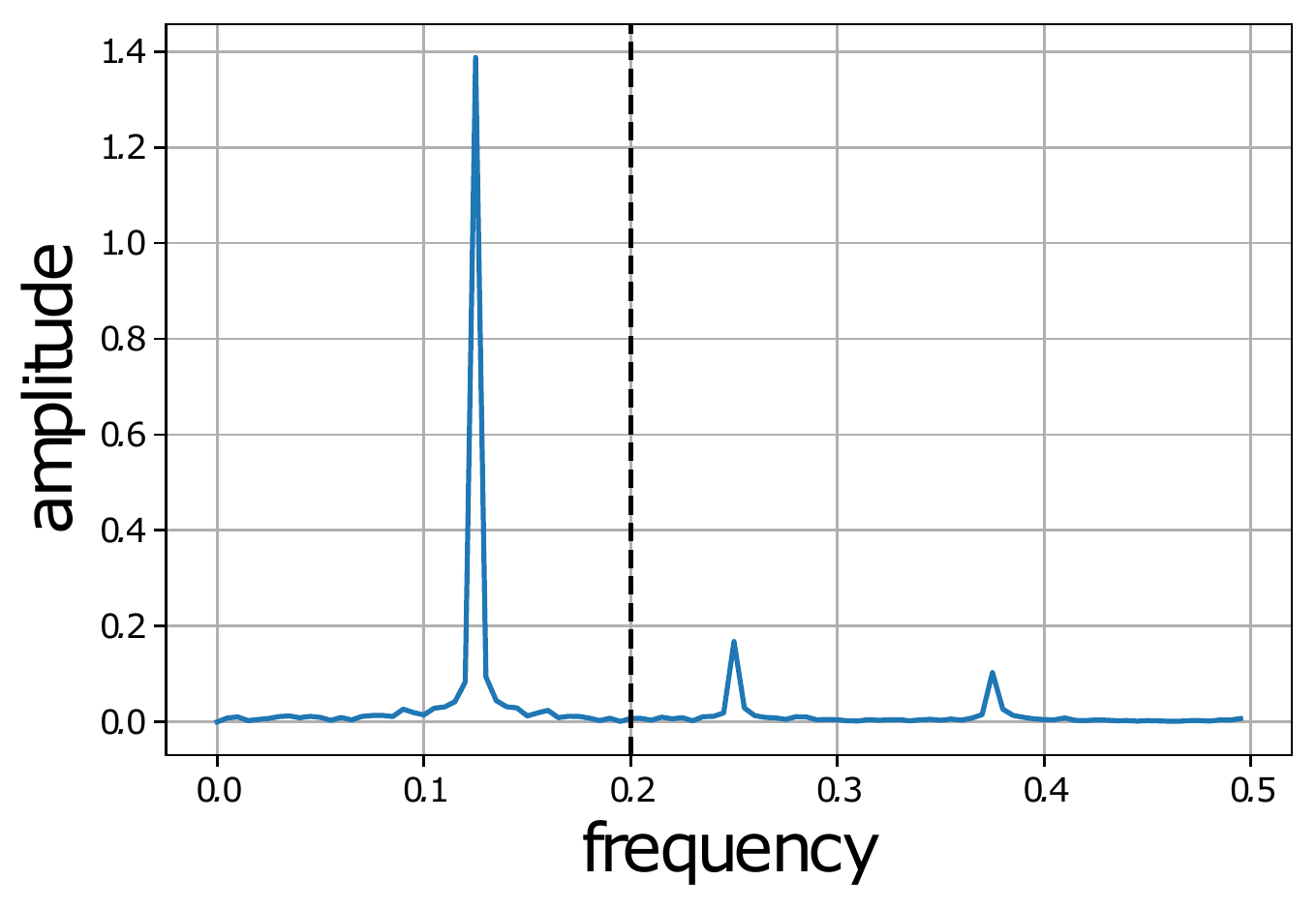}
		\caption{Double-torsion pendulum (iv)}\label{subfig:iv}
	\end{subfigure}		
	\caption{Fourier spectra for double-torsion pendulum with excitation ii) in Figure \ref{subfig:ii}, excitation (iii) in Figure \ref{subfig:iii} and iv) in Figure \ref{subfig:iv}.}
	\label{fig:freq_2}
\end{figure*}

\subsection{Double-mass spring system} \label{section:mass_model}
The double mass-spring system consists of two superposed sine waves. 
Here, we consider the following normalized system
\begin{equation}\label{eq:double-mass}
x(t)=1.28 \cos(2 \pi 0.115 t-7.7)+0.677 \cos(2\pi 0.57 t)-0.009.
\end{equation}
Training data is generated on an interval of 100 seconds with step size $dt=0.1$. 
Further, observation noise with variance $\sigma^2 = 0.1$ is added. 
\subsection{Van-der-Pol oscillator} \label{section:VDP_model}
Ground truth data are generated from the four-dimensional ground truth system 
\begin{equation} \label{eq:VDP}
\begin{aligned}
\begin{pmatrix}
\dot{x} \\
\dot{y} \\
\dot{u} \\
\dot{v}
\end{pmatrix}=
\begin{pmatrix}
y \\
-x + a(1-x^2)y+b u \\
v \\
- \omega^2 u
\end{pmatrix}.
\end{aligned}
\end{equation}
We assume that the first dimension $x$ referring to the position of the oscillator is observed.
The simulator refers to a standard Van-der-Pol oscillator, ignoring the external force 
\begin{equation} \label{eq:VDP_std}
\begin{aligned}
\begin{pmatrix}
\dot{x} \\
\dot{y} \\
\end{pmatrix}=
\begin{pmatrix}
y \\
-x + \tilde{a}(1-x^2)y \\
\end{pmatrix}
\end{aligned}.
\end{equation}
Here, we choose $a=5, b=80, \tilde{a}=3.81$. 
Data is simulated with a step size $dt=0.05$.

\subsection{Frequency domain}
In our experiments, a suitable cutoff frequency can be obtained by analyzing the frequency spectrum.
Here, we show the Fourier spectra of all methods and the chosen cutoff frequencies in order to gain insight into the system properties. 
For the systems that are used to train a hybrid model, we additionally depict the Fourier spectra of the simulator. 
Figure \ref{fig:freq_1} depicts the results for the double-mass spring system in Figure \ref{subfig:mass_f}, the Van-der-Pol oscillator in Figure \ref{subfig:VDP} and the drill-string system in Figure \ref{subfig:friction}. The results for the double-torsion pendulum with excitations (ii)-(iv) are shown in Figure \ref{fig:freq_2}.

\section{Additional experiments} \label{section:add_exp}
In this section, we provide additional experiments. 
We demonstrate the flexibility of our method in the purely learning-based scenario by applying different cutoff frequencies and downsampling rates. 
Furthermore, we provide additional plots for the results in Section 5. 

\subsection{Additional experiments} 
Here, we demonstrate the flexibility of our purely learning-based method. 
In particular, we compute the results for the double-mass spring system (cf. Sec. 5) with different cutoff frequencies and up/downsampling ratios. 
Since we consider data from a simulated double mass-spring system, we can derive the frequencies of the corresponding sine waves. 
Thus, the system is ideal in order to investigate the behavior of the method. 
In particular, the system consists of two superposed sine waves, one with frequency 0.57, one with frequency 0.115 (cf. Figure \ref{subfig:mass_f}).
Thus, the cutoff frequency should be chosen in this range.
Otherwise the predictions of one of the GRUs would be eliminated by the filter. 

Here, we choose cutoff frequencies $w=0.2 , w=0.1$, and $w=0.5$ with downsampling ratios $k=2$.
The results in Table \ref{t:RMSE_cutoff} demonstrate that the method is flexible with regards to varying cutoff frequencies. 

For the investigation of different downsampling rates, we choose $w=0.4$ as cutoff frequency. 
For the sampling frequency $f_{\textrm{sample}}$ it has to hold $2 w < f_{\textrm{sample}}$ in order to respect the Nyquist-Shannon theorem. 
With $w=0.4$ this yields $f_{\textrm{sample}}=0.8$. 
We choose $f_{\textrm{sample}}=1.0$ close to this frequency and thus a downsampling ratios $k=10$.
Further we investigate  $k=3, k=4$ and  $k=10$. 
Again, the results in Table \ref{t:RMSE_downsample} demonstrate that the method is flexible with regards to different downsampling ratios. 

\begin{table*}[ht]
	\caption{Total RMSEs (mean (std)) over 5 independent runs with varying cutoff frequencies.}
	\label{t:RMSE_cutoff}
	\centering 
	\begin{tabular}{rcc}
		\noalign{\smallskip} \hline \hline \noalign{\smallskip}
		cutoff frequency & split GRU (ours) & split GRU+HP (ours)  \\
		\hline
		0.2  & 0.149 (0.022) & 0.227 (0.043) \\
		0.4  &0.112 (0.025) & 0.116 (0.03)\\
		0.5  & 0.204 (0.022) & 0.195 (0.028)  \\
		\noalign{\smallskip} \hline \noalign{\smallskip}
	\end{tabular}\quad
\end{table*} 

\begin{table*}[ht]
	\caption{Total RMSEs (mean (std)) over 5 independent runs with varying downsampling rates.}
	\label{t:RMSE_downsample}
	\centering 
	\begin{tabular}{rcc}
		\noalign{\smallskip} \hline \hline \noalign{\smallskip}
		donwsampling rate k & split GRU (ours) & split GRU+HP (ours)  \\
		\hline
		2  & 0.12 (0.008) & 0.168 (0.03)\\
		5  &0.112 (0.025) & 0.116 (0.03)\\
		10 & 0.185 (0.023) & 0.187 (0.017)   \\
		\noalign{\smallskip} \hline \noalign{\smallskip}
	\end{tabular}\quad
\end{table*} 

\subsection{Additional plots}
\begin{figure*}[tb]
	\begin{subfigure}[htb]{0.33\textwidth}
		\centering
		\includegraphics[width= \textwidth]{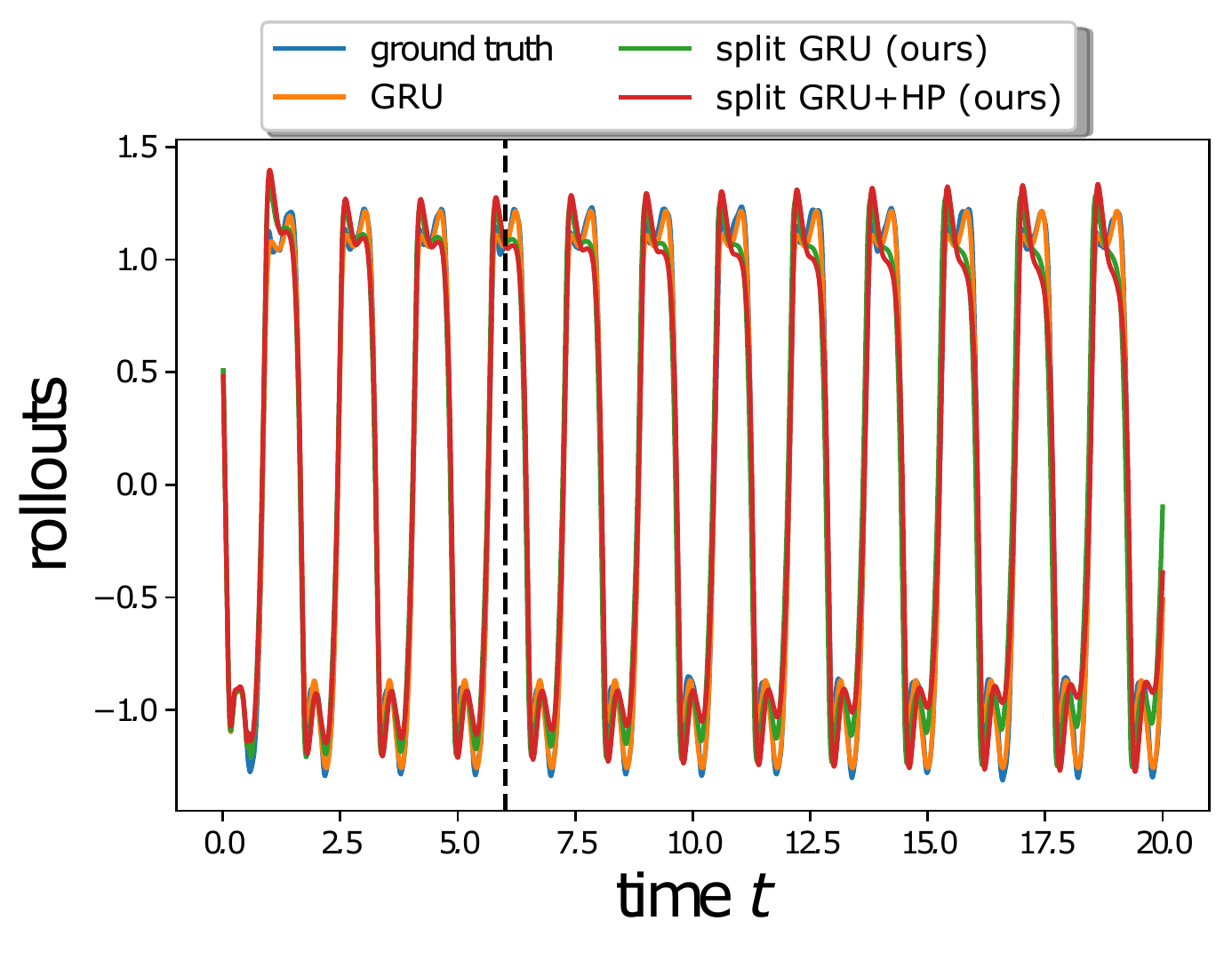}\label{subfig:friction_7}
		\caption{system iii)}\label{subfig:friction_7}
	\end{subfigure}	
	\begin{subfigure}[htb]{0.33\textwidth}
		\centering
		\includegraphics[width= \textwidth]{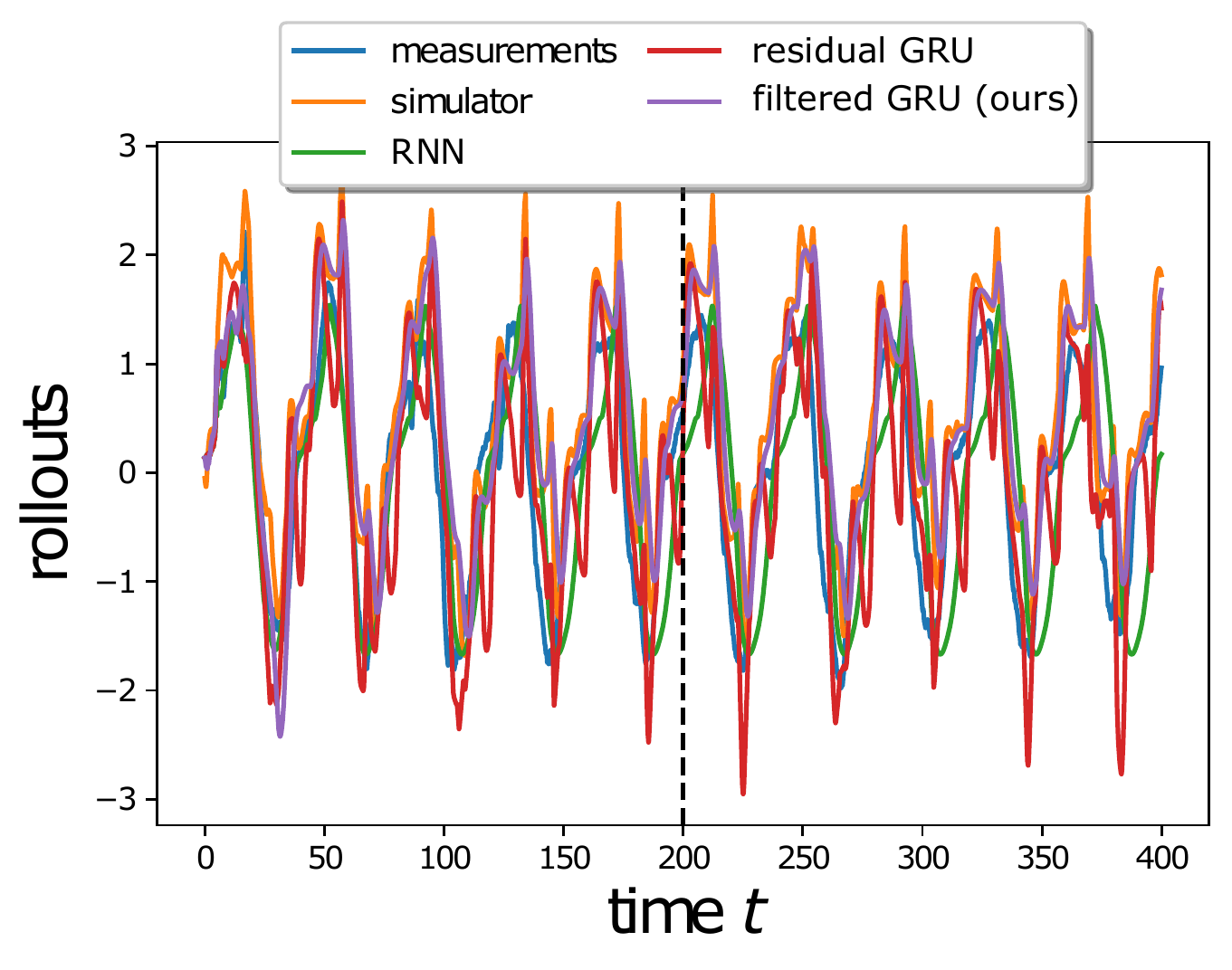}\label{subfig:friction_GRU}
		\caption{system vi) with GRU}\label{subfig:friction_GRU}
	\end{subfigure}	
	\begin{subfigure}[htb]{0.33\textwidth}
		\centering
		\includegraphics[width= \textwidth]{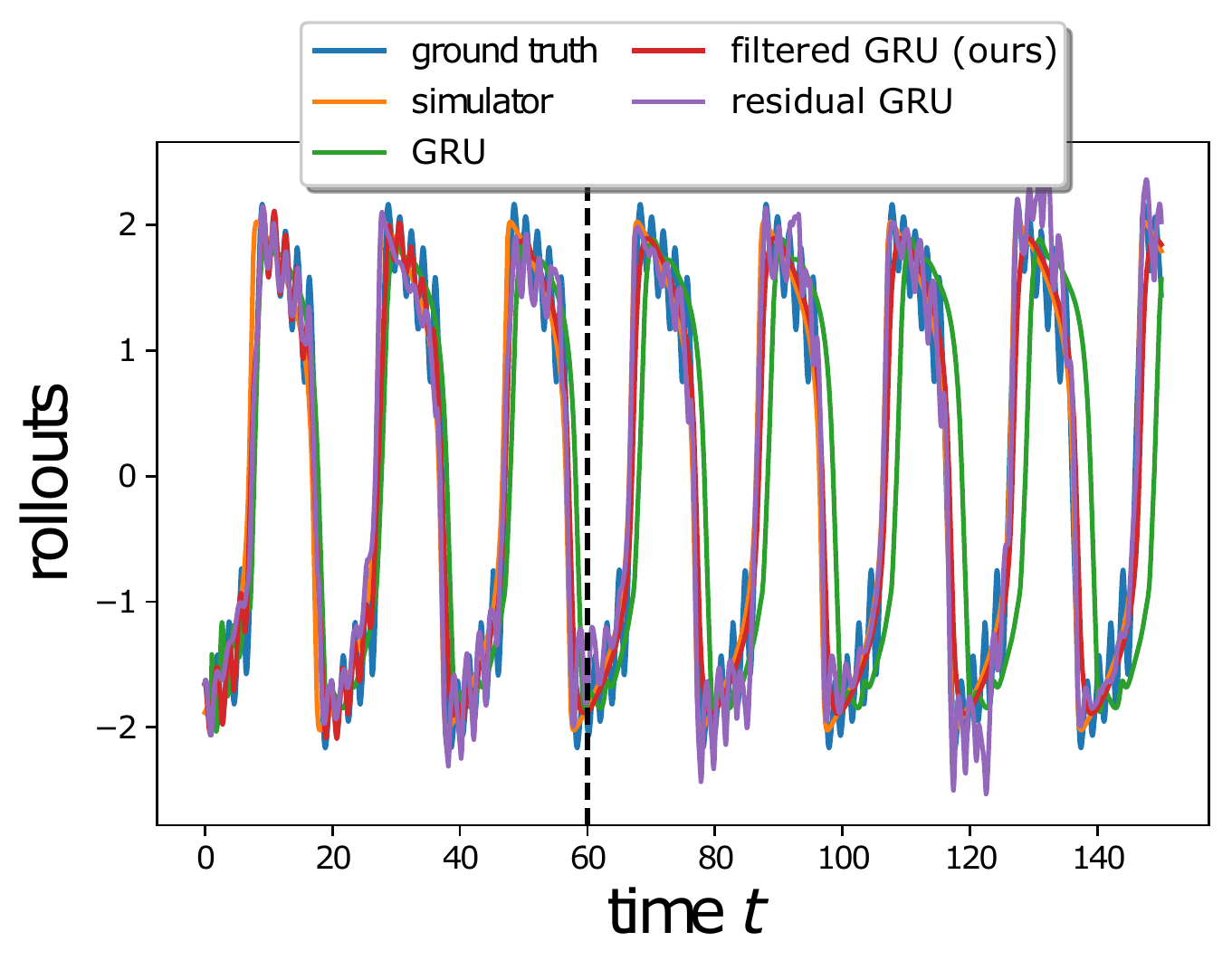}\label{subfig:VDP_GRU}
		\caption{system vi) with GRU}\label{subfig:VDP_GRU}
	\end{subfigure}	
	\caption{Shown are rollouts over time for the purely learning-based method on System (iii) in Figure \ref{subfig:friction_7}, for hybrid model with GRU on System (vi) in Figure \ref{subfig:friction_GRU} and for the Van-der-Pol oscillator with GRU on System (vi) in Figure \ref{subfig:VDP_GRU}. }\label{fig:add_plots}
\end{figure*}

\begin{figure*}[tb]
	\begin{subfigure}[htb]{0.33\textwidth}
		\centering
		\includegraphics[width= \textwidth]{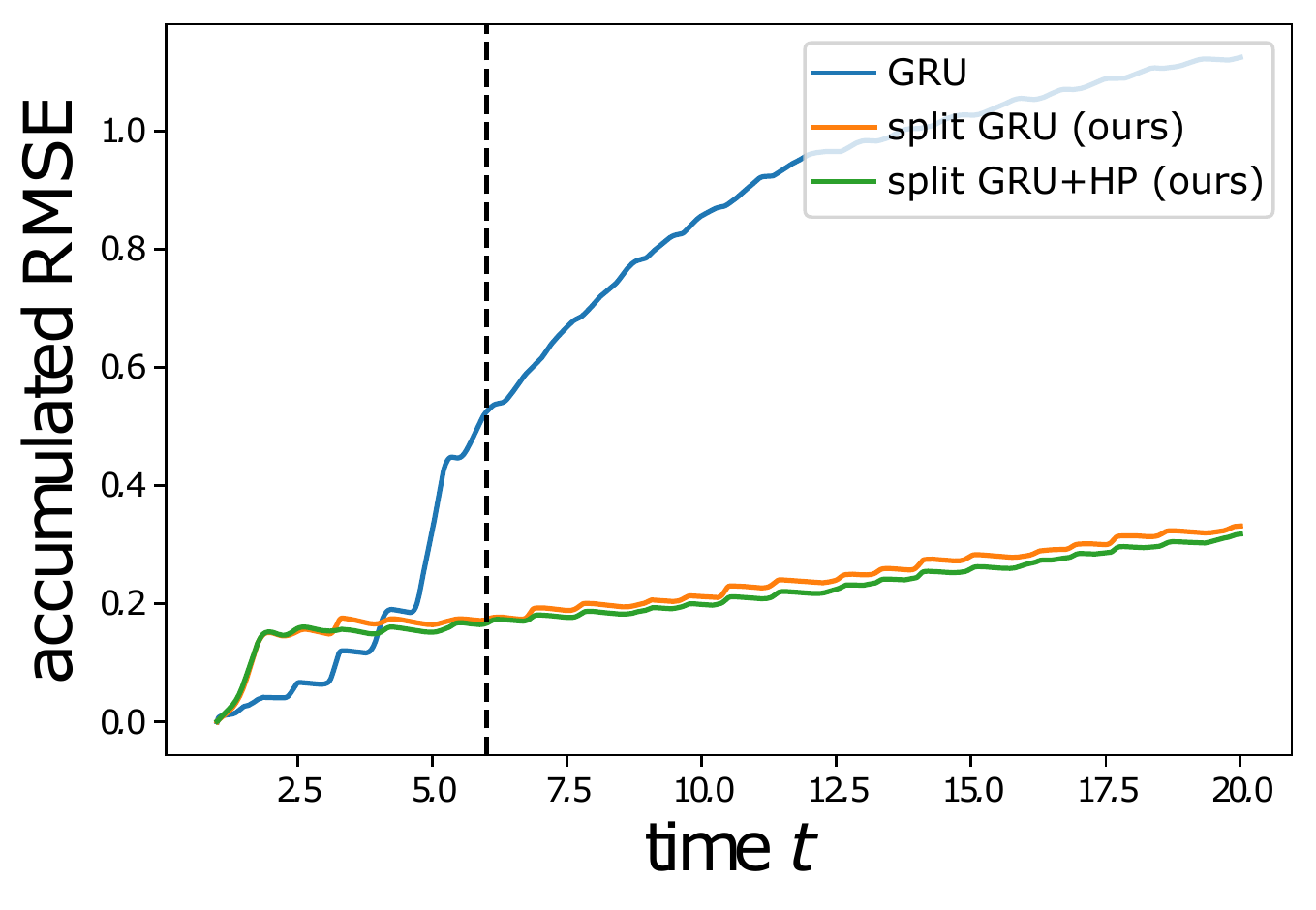}\label{subfig:loss_friction_5}
		\caption{system iii)}\label{subfig:loss_friction_5}
	\end{subfigure}	
	\begin{subfigure}[htb]{0.33\textwidth}
		\centering
		\includegraphics[width= \textwidth]{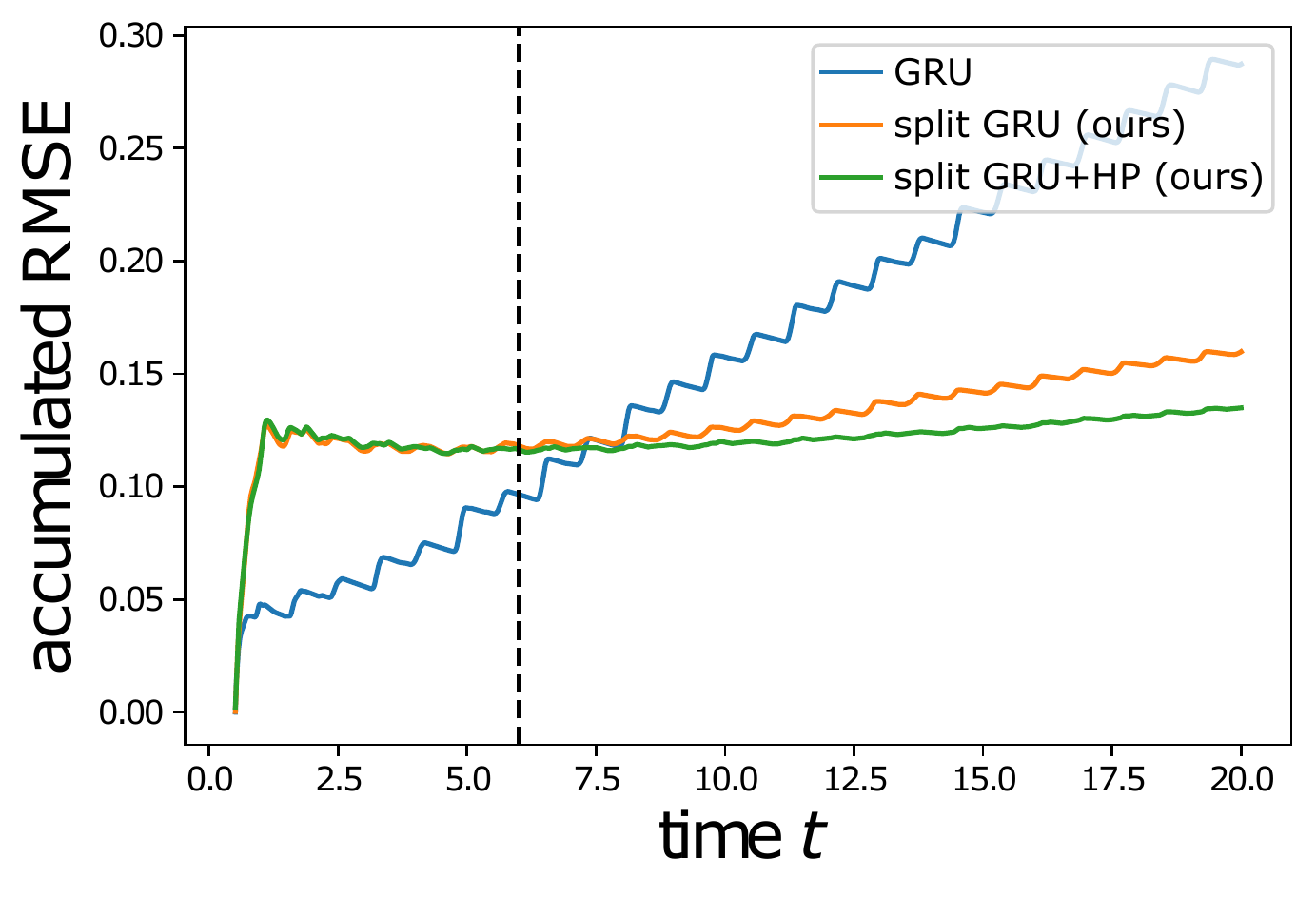}\label{subfig:loss_friction_7}
		\caption{system vi) with GRU}\label{subfig:loss_friction_7}
	\end{subfigure}	
	\begin{subfigure}[htb]{0.33\textwidth}
	\centering
	\includegraphics[width= \textwidth]{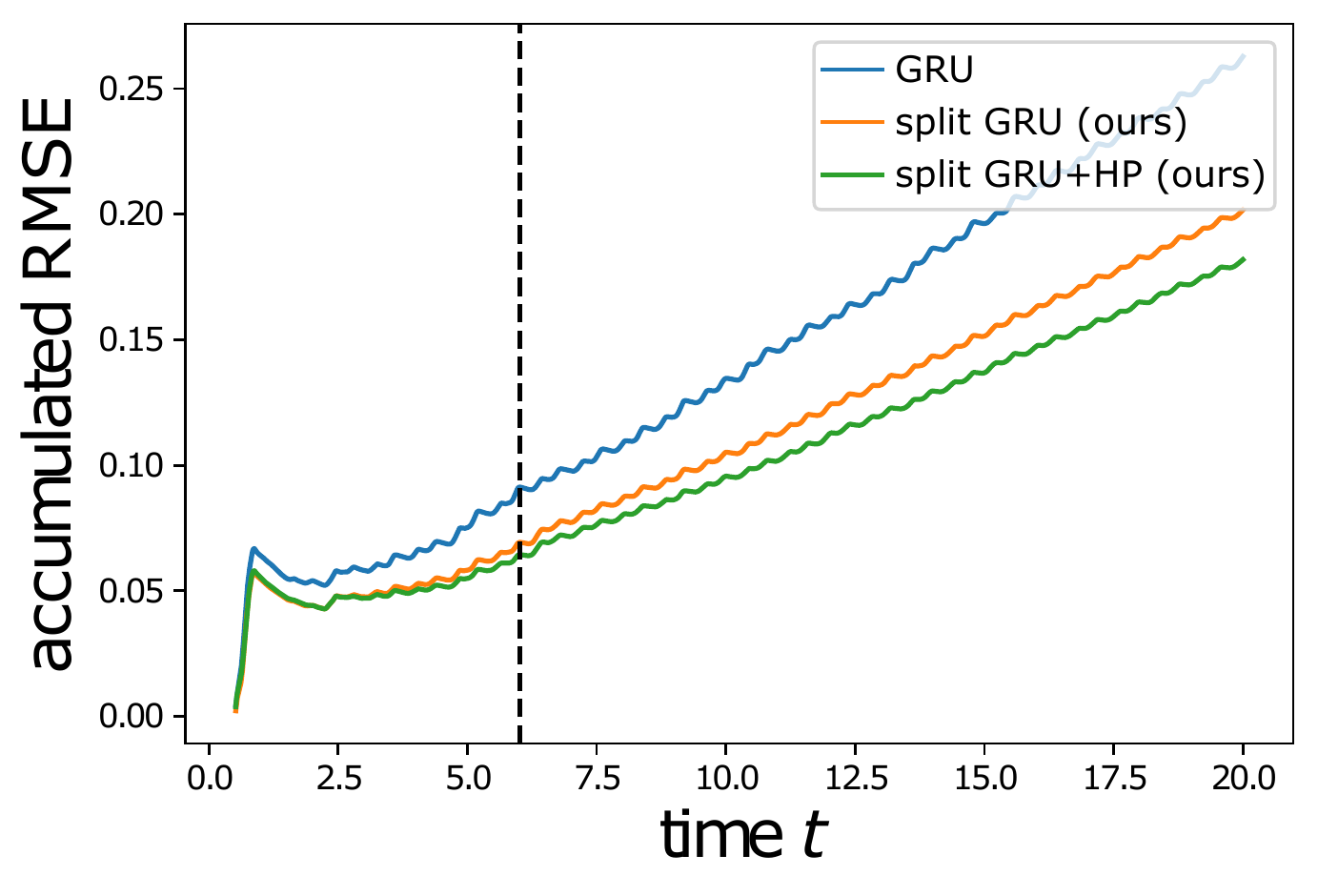}\label{subfig:loss_friction_8}
	\caption{system vi) with GRU}\label{subfig:loss_friction_8}
\end{subfigure}	
	\caption{Shown are absolute errors over time for the purely learning-based method on System (ii) in Figure \ref{subfig:loss_friction_5}, System (iii) in Figure \ref{subfig:loss_friction_7} and System (iv) in Figure \ref{subfig:loss_friction_8}. The results are computed after the warmup phase of the GRU.}\label{fig:lb}
\end{figure*}

\begin{figure*}[tb]
	\begin{subfigure}[htb]{0.49\textwidth}
		\centering
		\includegraphics[width= 0.6\textwidth]{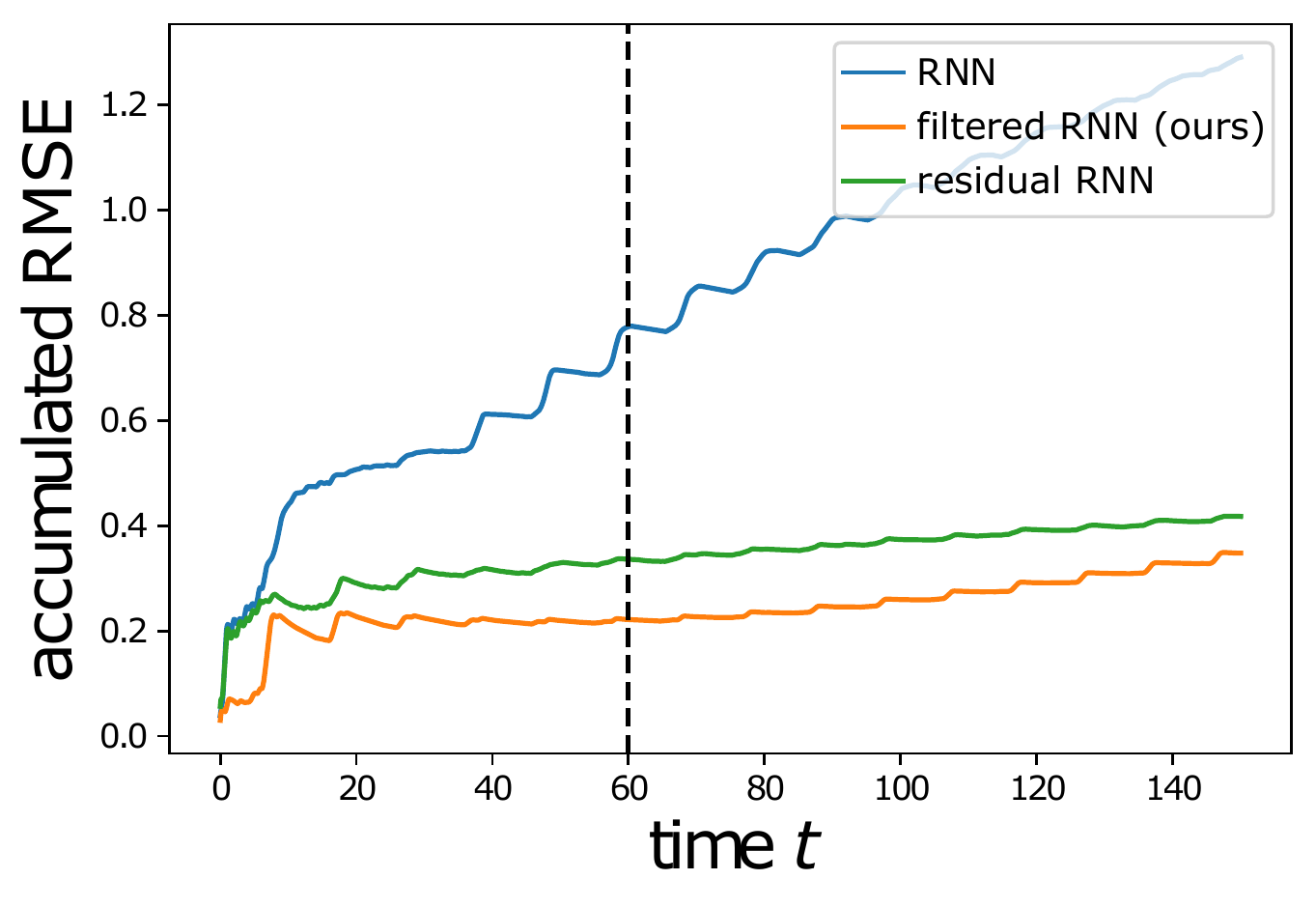}\label{subfig:VDPMLP}
		\caption{system v) with MLP}\label{subfig:VDPMLP}
	\end{subfigure}	
	\begin{subfigure}[htb]{0.49\textwidth}
		\centering
		\includegraphics[width= 0.6\textwidth]{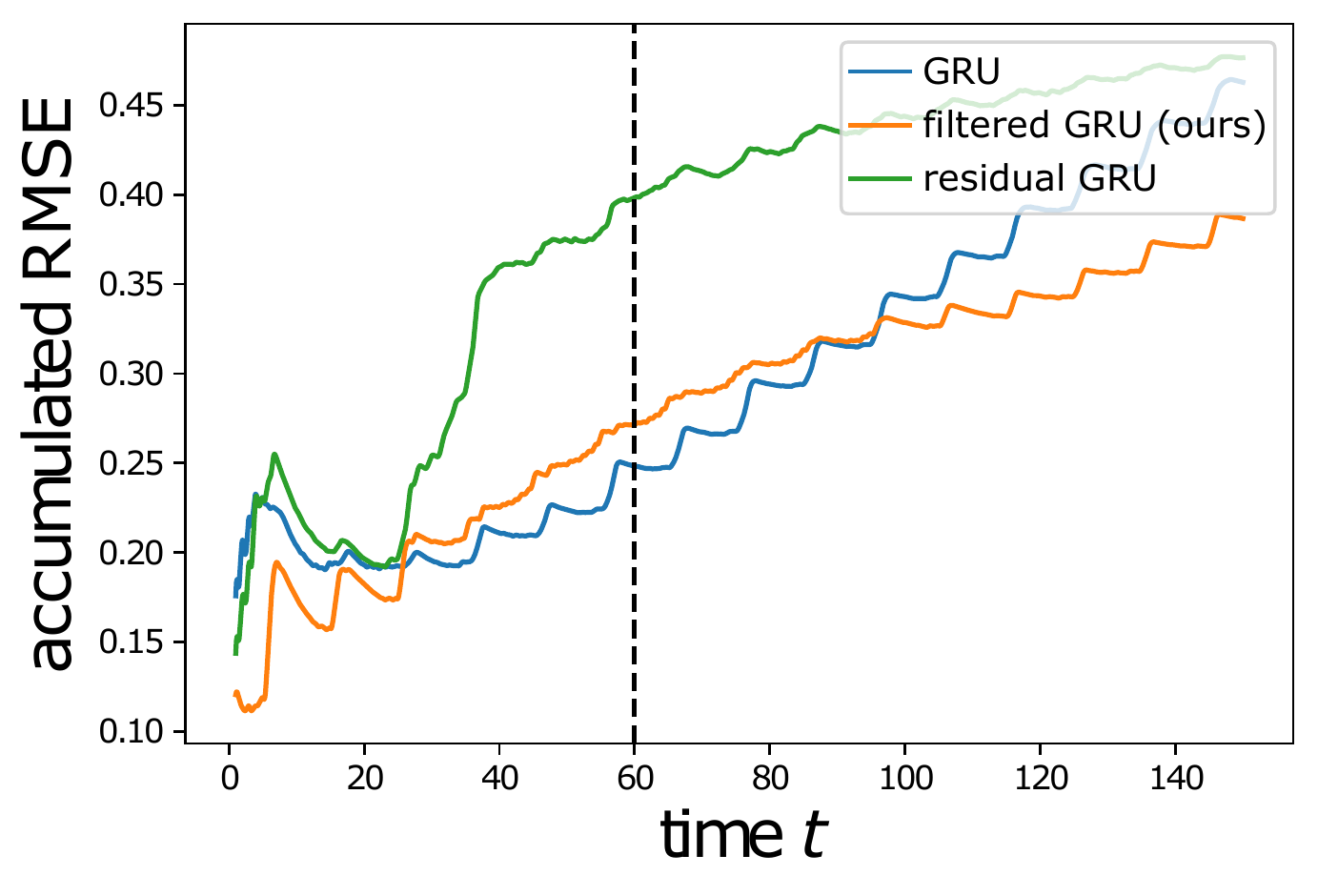}
		\caption{system v) with GRU}\label{subfig:VDPGRU}
	\end{subfigure}	
	\caption{Shown are absolute errors over time for the hybrid model on the Van-der-Pol oscillator (v) with MLP in Figure \ref{subfig:VDPMLP} and GRU in Figure \ref{subfig:VDPGRU}.
	The GRU and the corresponding residual models have difficulties to catch the oscillations in the beginning and therefore start with a high error.}\label{fig:hybridVDP}
\end{figure*}

\begin{figure*}[tb]
	\begin{subfigure}[htb]{0.49\textwidth}
		\centering
		\includegraphics[width= 0.6\textwidth]{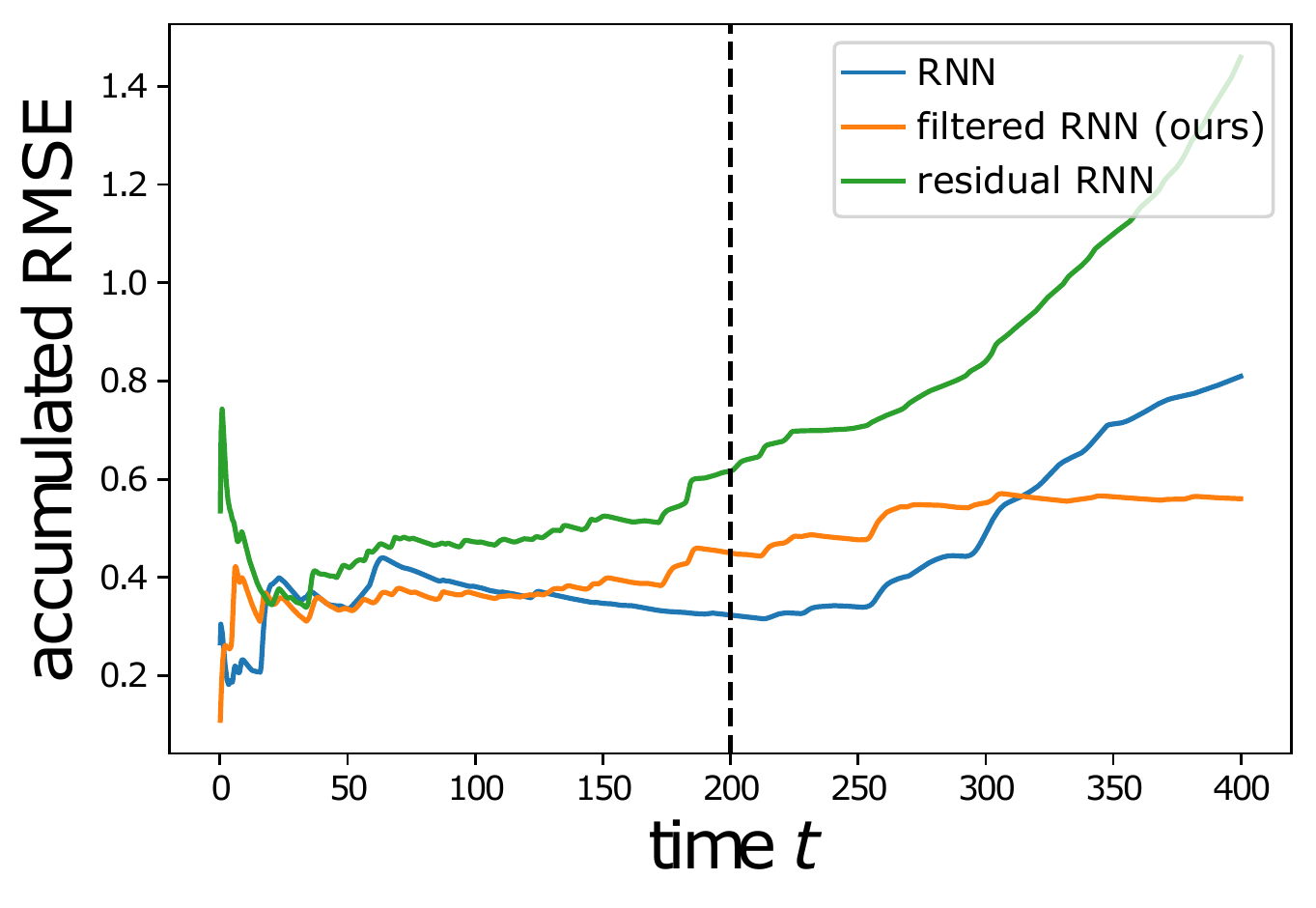}\label{subfig:frictionMLP}
		\caption{system vi) with MLP}\label{subfig:frictionMLP}
	\end{subfigure}	
	\begin{subfigure}[htb]{0.49\textwidth}
		\centering
		\includegraphics[width= 0.6\textwidth]{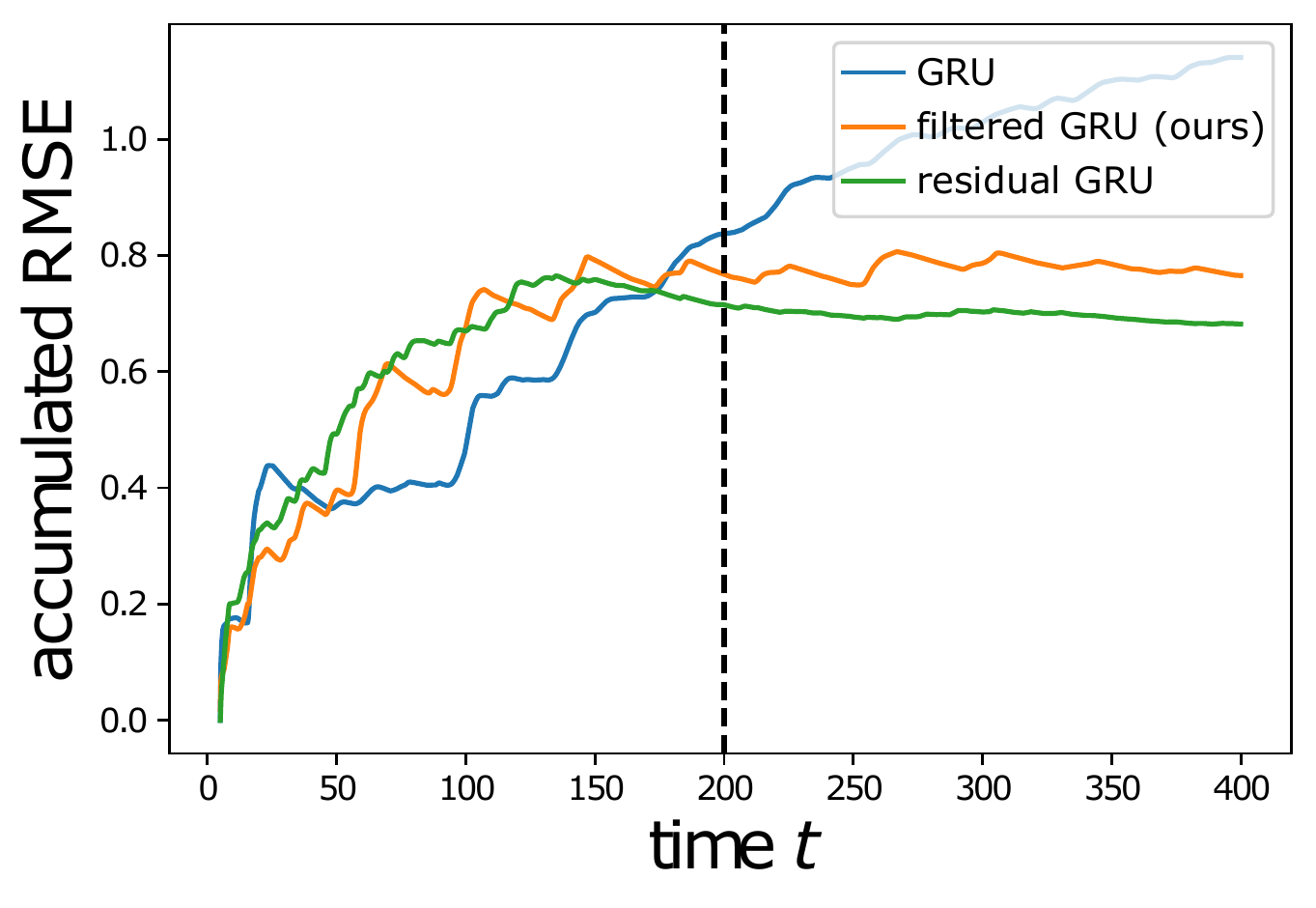}
		\caption{system vi) with GRU}\label{subfig:frictionGRU}
	\end{subfigure}
	\caption{Shown are absolute errors over time for the hybrid model for drill-string system (vi) with MLP in Figure \ref{subfig:frictionMLP} and GRU in Figure \ref{subfig:frictionGRU}. The residual model with MLP needs a few steps to converge in the beginning.}\label{fig:hybridfrition}	
\end{figure*}

We provide additional rollout plots for the results in Sec. 5 in Figure \ref{fig:add_plots}. Shown are plots for the double torsion pendulum with excitation iii) in Figure \ref{subfig:friction_7} and the friction model vi) with GRU in Figure \ref{subfig:friction_GRU}. 
Further, we provide the RMSEs over time for all systems in Figure \ref{fig:lb}, Figure \ref{fig:hybridVDP} and Figure \ref{fig:hybridfrition}.
For the GRU results, we compute the RMSEs after the warmup phases since the type of predictions differs from warmup phase to actual rollout.
For the MLP we compute the RMSEs along the whole trajectory, since no warmup phase is performed but the initial hidden state is estimated via the recognition model.

\newpage
\bibliography{bib}
%%%%%%%%%%%%%%%%%%%%%%%%%%%%%%%%%%%%%%%%%%%%%%%%%%%%%%%%%%%%
 
% Supplementary material has been moved to supplements, because website says so